\documentclass[acmtog,nonacm]{acmart}

\usepackage{booktabs} % For formal tables
\usepackage{multirow}
\usepackage{multirow}
\usepackage{threeparttable}
\usepackage{listings}
\lstdefinelanguage{json}{
  basicstyle=\ttfamily\small,
  showstringspaces=false,
  breaklines=true,
  literate=
    *{0}{{{\color{teal}0}}}{1}
     {1}{{{\color{teal}1}}}{1}
     {2}{{{\color{teal}2}}}{1}
     {3}{{{\color{teal}3}}}{1}
     {4}{{{\color{teal}4}}}{1}
     {5}{{{\color{teal}5}}}{1}
     {6}{{{\color{teal}6}}}{1}
     {7}{{{\color{teal}7}}}{1}
     {8}{{{\color{teal}8}}}{1}
     {9}{{{\color{teal}9}}}{1}
     {:}{{{\color{black}{:}}}}{1}
     {,}{{{\color{black}{,}}}}{1}
     {\{}{{{\color{black}{\{}}}}{1}
     {\}}{{{\color{black}{\}}}}}{1}
     {[}{{{\color{black}{[}}}}{1}
     {]}{{{\color{black}{]}}}}{1},
  morestring=[b]",
  stringstyle=\color{brown},
}
\lstdefinestyle{python}{
  language=Python,
  basicstyle=\ttfamily\small,
  showstringspaces=false,
  breaklines=true,
  keywordstyle=\color{blue!70!black},
  keywordstyle=[2]\color{teal},
  morekeywords=[2]{None, True, False},
  stringstyle=\color{brown},
  commentstyle=\color{gray}\itshape,
  emphstyle=\color{purple},
  emph={def, return, self},
  literate=
    *{0}{{{\color{teal}0}}}{1}
     {1}{{{\color{teal}1}}}{1}
     {2}{{{\color{teal}2}}}{1}
     {3}{{{\color{teal}3}}}{1}
     {4}{{{\color{teal}4}}}{1}
     {5}{{{\color{teal}5}}}{1}
     {6}{{{\color{teal}6}}}{1}
     {7}{{{\color{teal}7}}}{1}
     {8}{{{\color{teal}8}}}{1}
     {9}{{{\color{teal}9}}}{1}
     {(}{{{\color{black}{(}}}}{1}
     {)}{{{\color{black}{)}}}}  {1}
     {:}{{{\color{black}{:}}}}{1}
     {,}{{{\color{black}{,}}}}{1},
}

\lstdefinelanguage{markdown}{
  basicstyle=\ttfamily\small,
  showstringspaces=false,
  breaklines=true,
  morecomment=[l]{\#},
  commentstyle=\color{blue!70!black}\bfseries,
  morestring=[b]`,
  stringstyle=\color{brown}\ttfamily,
  literate=
    {**}{{{\color{purple}\textbf{**}}}}{2}
    {*} {{{\color{teal}{*}}}}{1}
    {-} {{{\color{gray}{-}}}}{1}
    {>} {{{\color{gray}{>}}}}{1},
}
\usepackage{url}
\usepackage[table,dvipsnames,x11names]{xcolor}
\usepackage[ruled]{algorithm2e} % For algorithms

\SetAlFnt{\small}
\SetAlCapFnt{\small}
\SetAlCapNameFnt{\small}
\SetAlCapHSkip{0pt}

\acmJournal{TOG}
\begin{document}
% Title portion
\title{\textsc{\textsc{Function2Scene}}: 3D Indoor Scene Layout from Functional Specifications}

% DO NOT ENTER AUTHOR INFORMATION FOR ANONYMOUS TECHNICAL PAPER SUBMISSIONS TO SIGGRAPH 2019!
\author{Ruiqi Wang}
\orcid{0009-0000-3379-6103}
\affiliation{%
 \institution{Simon Fraser University}
 \city{Burnaby}
 \state{BC}
 \country{Canada}
 }
\email{ruiqi_w@sfu.ca}
\author{Qimin Chen}
\affiliation{%
 \institution{Simon Fraser University}
 \city{Burnaby}
 \country{Canada}
}
\email{qca143@sfu.ca}
\author{Daniel Ritchie}
\affiliation{%
\institution{Brown University}
\city{Providence}
\state{Rhode Island}
\country{USA}}
\email{daniel_ritchie@brown.edu}
\author{Angel X. Chang}
\affiliation{%
 \institution{Simon Fraser University}
 \city{Burnaby}
 \country{Canada}
}
\email{angelx@sfu.ca}
\author{Manolis Savva}
\affiliation{%
 \institution{Simon Fraser University}
 \city{Burnaby}
 \country{Canada}
}
\email{manolis.savva@gmail.com}
\author{Kai Wang}
\affiliation{%
 \institution{Simon Fraser University}
 \city{Burnaby}
 \country{Canada}
}
\affiliation{%
 \institution{ShanghaiTech University}
 \country{China}}
\email{kwang.ether@gmail.com}
\author{Hao Zhang}
\affiliation{%
 \institution{Simon Fraser University}
 \city{Burnaby}
 \country{Canada}
}
\email{haoz@sfu.ca}

\begin{abstract}
\if 0
Most 3D indoor scene synthesis methods focus on visual appearances and/or rely on object-centric prompts that specify what to place.
In practice, people use the indoor environments, and care about what functionalities the rooms support---what occupants, and what activities.
We study layout generation from such a functional perspective, and propose \textsc{Function2Scene}, a functional scene synthesis framework that parses a functional design specification into personas and activities, derives a tailored set of design constraints from a taxonomy of 17 criteria spanning spatial, ergonomic, activity, and environmental categories, grounded in interior-design literature.
We use these constraints to iteratively refine a generated layout with a LLM that has access to a wide variety tools designed for verifying such constraints.
We demonstrate that layouts produced by \textsc{Function2Scene} are heavily preferred over three recent baselines in a crowd-sourced two-alternative forced choice study.
\fi
Most text-driven 3D indoor scene synthesis methods generate rooms from object-centric prompts, asking what furniture should be placed rather than how the space is used. Yet in real interior design, a layout is judged by how well it supports its occupants, e.g., their activities and physical needs. We introduce \textsc{Function2Scene}, a framework for generating 3D indoor layouts from \emph{functional} specifications, i.e., natural-language design briefs describing who will use a room and what they need to do there. Given such a specification, our system parses occupant personas and activities, derives a customized set of functional design constraints from a taxonomy of 17 criteria spanning spatial, ergonomic, activity, and environmental considerations, and uses these constraints to guide layout generation. Rather than relying on an LLM to directly produce a final scene, \textsc{Function2Scene} performs iterative evaluation and refinement through a tool-augmented check-and-repair loop, combining geometric measurements, LLM-based contextual reasoning, and VLM-based visual assessment. Experiments on 30 professionally written interior-design cases show that \textsc{Function2Scene} produces layouts that better satisfy functional requirements than recent LLM-based scene synthesis baselines, with our results preferred in 94.3\% of pairwise comparisons. Our work reframes text-driven indoor scene synthesis from placing plausible objects to designing spaces that support human use.

\end{abstract}

%     <concept>
%     <concept_id>10010147.10010178.10010224.10010225.10010227</concept_id>
%     <concept_desc>Computing methodologies~Scene understanding</concept_desc>
%     <concept_significance>500</concept_significance>
%     </concept>
%     <concept>
%     <concept_id>10010147.10010257</concept_id>
%     <concept_desc>Computing methodologies~Machine learning</concept_desc>
%     <concept_significance>300</concept_significance>
%     </concept>
%  </ccs2012>

% \ccsdesc[500]{Computing methodologies~Scene understanding}
% \ccsdesc[300]{Computing methodologies~Machine learning}


% \keywords{Scene synthesis, layout generation, large languge models, functionality}
\begin{teaserfigure}
    \centering
    \includegraphics[width=0.99\linewidth]{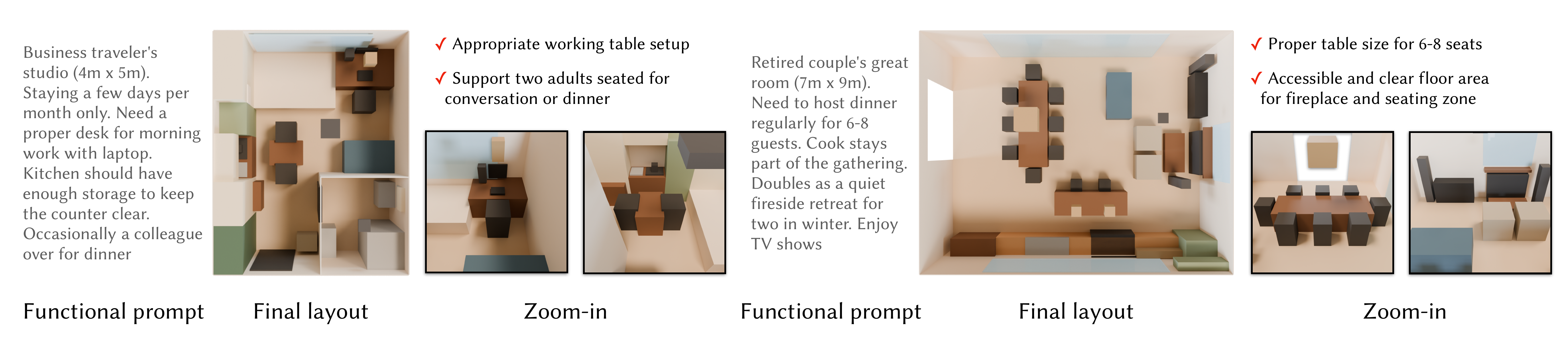}
    \vspace{-1em}
    \caption{We present \textsc{\textsc{Function2Scene}}, a framework for generating 3D indoor layouts from functional specifications. Given a detailed functional specification, our method decompose them into functional design constraints, which are then used to iteratively evaluate and refine a generated scene. Please refer to the supplementary material for the full input prompt and more detailed visualizations.}
    \Description{This is the teaser figure for the article.}
    \label{fig:teaser}
\end{teaserfigure}

\maketitle

\section{Introduction}
\label{sec:intro}

%A furnished room is not only a collection of objects, but a proposal for how a space should be used.
%Furniture layout determines how people circulate, where they sit and look, what they can reach, which activities can happen, and how daylight and noise affect them.
%A room can be visually and semantically plausible, yet still be a poor design: a desk facing direct window glare, a sofa blocking the path to the door, a wardrobe being unreachable by a senior who uses it, or a child's play area being hidden from their caregiver.
%Function is not a secondary attribute, it is one of the main reasons behind the layout.

A furnished room is not only a collection of objects, but a proposal for how a space should be used. Furniture layout determines how people circulate, where they sit and look, what they can reach, and which activities can happen. % and how daylight and noise affect them. 
A room can be visually and semantically plausible, yet still be a poor design: a desk facing direct window glare, a sofa blocking the path to the door, a wardrobe unreachable by a senior who uses it, or a child's play area hidden from their caregiver. Function is not a secondary attribute --- it is one of the primary reasons behind any layout decision.

The computer graphics community has long recognized and studied this.
Early scene generation works encoded design guidelines, such as clearance, grouping, alignment, accessibility, as explicit cost functions~\cite{merrell2011interactive, yu2011make}, but authoring such constraints required significant expertise and effort, making them hard to adapt to different scenarios.
To address this limitation, the field pivoted to data-driven approaches, starting with pairwise spatial priors~\cite{yu2011make, fisher2012example}, and advancing through the deep learning era with increasingly expressive learned priors with convolutional neural networks~\cite{wang2018deep}, autoregressive transformers~\cite{paschalidou2021atiss}, and denoising diffusion~\cite{tang2024diffuscene, zhai2023commonscenes}.
These work capture increasingly rich statistical patterns, but in doing so, make functional design knowledge more and more implicit and uninterpretable. 

Entering the foundation model era, LLM-based systems~\cite{feng2023layoutgpt, yang2024holodeck, ccelen2024idesign, feng2025casagpt, sun2025layoutvlm} enabled flexible text conditioned scene synthesis.
More recently, agentic pipelines also enable iterative scene updates to further improve scene quality~\cite{he2026sceneorchestra, luo2026sceneassistant, xia2026sage, zhao2026scenerevis}.
Leveraging foundational priors, these systems support more open-ended scene generation. 
However, they largely inherit the learning era's ``implicit" approach: an LLM directly generates objects, transforms relations, and even in agentic setups, the result is mostly checked only for visual quality and physical plausibility.
Consistent with their goals, most LLM-based methods consume detailed object-centric prompts, for example, ``a bedroom with a queen bed, two nightstands, and a dresser.''
They are not targeting more ``functionality"-oriented prompts, for example, ``a bedroom for a couple where one partner reads late while the other sleeps early.'' 
Yet, the latter describes a far more common starting point in real design practices~\cite{kilmer2024designing, panero1962anatomy}.
The irony is that LLMs are well suited to the two tasks that limited classical rule-based layout generation in the first place: parsing open-ended functional descriptions, and optimizing functional criteria that are cost prohibitive to manually specify.
 
We study 3D indoor scene layout from \emph{functional specifications}: natural-language design briefs that describe who will use a space, what they will do in it, and what constraints their needs impose.
Such briefs are closer to an interior design program than to a short text-to-scene command.
For the scope of this work, we use professionally written room descriptions adapted from sources such as \emph{Architectural Digest}.
%We leave the problem of assisting non-expert users to arrive at such descriptions for future work.
This problem formulation introduces unique challenges, as functional specifications are inherently high-level: instead of directly specifying objects and layout, they impose diverse and heterogeneous constraints over the indoor space: spatial constraints, ergonomic rules, activity patterns, environmental contexts, etc.
Consequently, LLMs tasked to directly generate scenes from such functional specification not only struggle with producing scenes that meet the functional demands, but also frequently fail to produce meaningful scenes, without the availability of more explicitly worded prompts.

To address these challenges, we introduce \emph{Function2Scene}, a framework that empowers LLMs for such functional scenarios.
Given a functional specification, we begin by analyzing the ``personas," i.e., who the occupants are and what specific needs they have, and the ``activities," i.e., what the occupants do in the space.
We then generate a detailed scene description, as well as a set of design guidelines drawn from a taxonomy of 17~criteria organized into four categories, Spatial, Ergonomic, Activity, and Environmental, grounded in interior-design literature~\cite{kilmer2024designing, panero1962anatomy}, that conforms to the personas and their activities.
This revisits the classical idea of codifying design rules for layout generation~\cite{merrell2011interactive, yu2011make, leimer2022layoutenhancer}, but makes the selected rules customized to the specific functional specifications, a customization that is only possible thanks to the foundational knowledge of LLMs.
Finally, we generate an initial layout and iteratively refine it through a check-and-repair loop to make the layout conform to the parsed design guidelines.

%Our functionality-aware scene generation pipeline is able to generate diverse types of scenes following a variety of real-world functional specifications, that are universally preferred over baselines and ablations in a crowd-sourced two-alternative forced choice (2AFC) perceptual study.

Our functionality-aware pipeline generates diverse scenes from real-world functional specifications, with results preferred over all baselines and ablations in 94.3\% of pairwise comparisons in a crowd-sourced two-alternative forced choice (2AFC) perceptual study.

In summary, our contributions are:
\vspace{-0.5em}
\begin{itemize}
    %\item A novel angle for 3D indoor layout generation that emphasis functionality over appearance, exposing failure modes not directly targeted by existing layout generators.
    \item A functionality-first framing for 3D indoor layout generation, shifting the input from object-centric prompts to functional specifications, and exposing failure modes not addressed by existing scene synthesis methods.
    \item A design constraint taxonomy (Spatial, Ergonomic, Activity, Environmental) rooted in interior-design literature, together with an LLM-driven method for automatically customizing constraints to specific occupant personas and activities.
    \item An iterative layout generation framework that combines geometric measurements, LLM-based reasoning, and VLM-based visual assessment in a tool-augmented check-and-repair loop to generate high-quality, functionally valid layouts from functional specifications.
\end{itemize}

\section{Related Works}
\label{sec:related}

\paragraph{Scene synthesis pre-LLMs.}
Indoor scene synthesis has long been studied in computer graphics.
Early works leaned heavily on pre-specified design principles~\cite{merrell2011interactive}, simple statistical relationships~\cite{yu2011make} and hand-written programs~\cite{yeh2012synthesizing}, which all require heavy manual effort, and is not comprehensive enough to handle open-ended settings, even when massively scaled up~\cite{deitke2022procthor,raistrick2024infinigen}.
As a result, the field gradually shifted towards data-driven approaches~\cite{fisher2012example, kermani2016learning, liang2017automatic}, and become dominated by deep learning based approaches~\cite{wang2018deep, li2019grains, zhang2020deep, ritchie2019fast, zhou2019scenegraphnet, wang2019planit, paschalidou2021atiss, tang2024diffuscene, lin2024instructscene, bai2025freescene}.
Such learning-based approaches require minimal manual effort, but generally do not learn explicit design principles, making them hard to adapt for our settings.
LayoutEnhancer~\cite{leimer2022layoutenhancer} attempts to address this limitation by directly injecting ergonomic principles into learned generative models, yet they also involve manual rule authoring that is hard to scale.
A parallel line of work has argued that scene plausibility should be grounded in human use.
SceneGrok~\cite{savva2014scenegrok} predicted where actions occur from observations, activity-centric synthesis~\cite{fisher2015activity} used such predictions to guide scene generation, and PiGraphs~\cite{savva2016pigraphs} learned joint models of human pose and scene geometry from interactions.
Subsequent work conditioned layout on activity types~\cite{fu2017adaptive, ma2016action, qi2018human} or optimized for human-aware navigation~\cite{sun2023haisor}.
These methods established that activities can usefully guide object selection and placement, but they map activities directly to arrangements rather than to the underlying design constraints that determine whether a layout actually supports the activity.
 
\begin{figure*}[t]
    \centering
    \includegraphics[width=\linewidth]{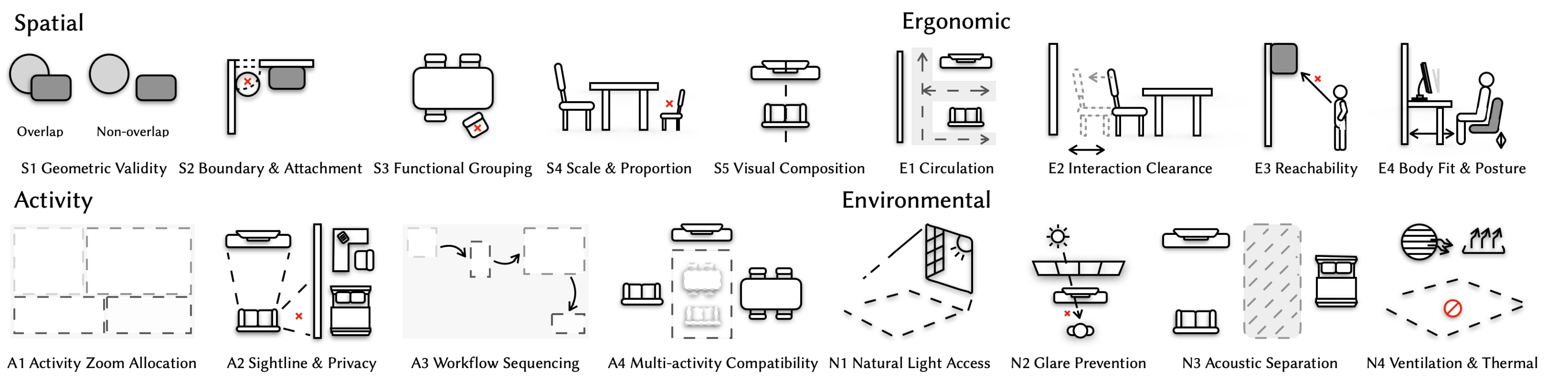}
    \caption{\textbf{Constraints Taxonomy.} We organize interior design constraints into four categories: Spatial (S1--S5), Ergonomic (E1--E4), Activity (A1--A4), and Environmental (N1--N4), each illustrated with representative examples of how they shape furniture placement in a typical room layout.}
    \label{fig:constraints}
    \vspace{-1em}
\end{figure*}

\paragraph{LLM-based scene synthesis.}
While text-to-scene generation long predates~\cite{chang2015text, ma2018language} large language models, the rise of LLMs transformed the scale and flexibility of what language-conditioned systems can produce.
LayoutGPT~\cite{feng2023layoutgpt} showed that LLMs can directly predict object coordinates from open-vocabulary prompts, Holodeck~\cite{yang2024holodeck} scaled this to full embodied environments with constraint satisfaction, and I-Design~\cite{ccelen2024idesign} added personalization from user preferences.
Since then, the field has expanded rapidly: hierarchical and structured representations~\cite{wang2025hlg, zhou2025roomcraft, zhang2025scene, ocal2024sceneteller, pun2025hsm, tam2024smc, wu2025diorama}, deeper spatial reasoning via chain-of-thought~\cite{ran2025direct} or VLM-guided search~\cite{deng2025global, berdoz2025text}, design-aware placement~\cite{feng2025casagpt, yang2025optiscene, bucher2025respace, guptainterioragent}, and agentic pipelines with iterative self-correction~\cite{yang2025sceneweaver, luo2026sceneassistant, liu2025agentic, he2026sceneorchestra, sun2025generalist, xia2026sage}.
Yet throughout this progression, the input prompt is predominantly about objects, relation and coordinates.
Even systems that incorporate richer signals---personalized preferences~\cite{ccelen2024idesign, yang2024llplace}, physical interactability~\cite{yang2024physcene}, or disentangled semantic-physical refinement~\cite{gao2025disco, pan2025metaspatial}---optimize primarily for spatial plausibility and visual coherence rather than for the functional, ergonomic, and environmental criteria that determine whether a room actually supports the activities it was designed for.
 
\paragraph{Scene evaluation and optimization.}
How a generated layout is evaluated shapes what it can become.
Current evaluation frameworks assess geometric validity, semantic coherence, navigability, and collision avoidance~\cite{tam2026sceneeval, hwangbo2025lego}, and iterative optimization loops---VLM-based feedback~\cite{asano2025geometry, jiang2026hog}, multi-turn RL~\cite{zhao2026scenerevis}, differentiable VLM optimization~\cite{sun2025layoutvlm}, and VL-guided editing~\cite{bian2025holodeck}---have made it possible to progressively refine layouts after initial generation~\cite{feng2026repurposing}.
Recent work has begun extending evaluation toward functional affordance grounding~\cite{maillard2026sceneteract}, but existing frameworks still lack systematic coverage of the human-centered criteria---ergonomic fit, activity support, environmental comfort---that govern how people actually use indoor spaces.
Our framework addresses this by employing typed verification tools---geometric checks for measurable spatial properties, LLM queries for contextual semantic judgments, and VLM assessments for holistic visual quality---organized across four constraint categories to evaluate and iteratively refine layouts against functional design criteria.

\section{Design Constraints}
\label{sec:constraints}
Professional interior designers compose layouts by following a rich set of principles and guidelines drawn from both design literature and practical experience~\cite{kilmer2024designing, panero1962anatomy}. 
Prior work has formalized such guidelines as optimization criteria for automated layout generation~\cite{merrell2011interactive}, yet these criteria are applied uniformly regardless of who occupies the space or what they do in it. 

\begin{table*}[t]
    \centering
    \footnotesize
    \caption{\textbf{Evaluation tools setup.} Each constraint is verified by one or more tools, color-coded by type: \colorbox{YellowGreen!40}{numeric/geometric} tools compute quantitative measures directly from scene geometry, \colorbox{Apricot!40}{LLM query} tools leverage language model reasoning over structured scene data, and \colorbox{RoyalBlue!30}{VLM} tools interpret rendered images. Tier indicates evaluation priority, where lower-tier constraints are verified first as prerequisites for higher-tier ones.}
    \vspace{-1em}
    \begin{tabular}{clllc}
    \toprule
    Category & Constraints & Tools & What it checks (Return) & Tier \\
    \midrule
     \multirow{16}{*}{\rotatebox[origin=c]{90}{\parbox[c]{1cm}{\centering Spatial}}} & \multirow{2}{*}{S1 Geometry Validity} & \colorbox{YellowGreen!40}{boundary\_check()} & within-wall containment (bool); & \multirow{2}{*}{T1} \\
     & & \colorbox{YellowGreen!40}{bbox\_collison()} & pairwise overlap ratio (\%)\\
     \cmidrule{2-5}
     & \multirow{2}{*}{S2 Boundary \& Attachment} & \colorbox{YellowGreen!40}{contact\_check()} & floor/ceiling/wall attachment (bool);  & \multirow{2}{*}{T1}\\
     & & \colorbox{YellowGreen!40}{wall\_angle\_check()} & object-to-wall angle (degrees)\\
     \cmidrule{2-5}
     & \multirow{2}{*}{S3 Spatial Relationships} & \colorbox{YellowGreen!40}{object\_exist()} & object presence (bool);  & \multirow{2}{*}{T2}\\
     & & \colorbox{YellowGreen!40}{object\_info()} & object size, location and orientation data ($l,w,h$, $x,y,z$, facing)\\
     \cmidrule{2-5}
     & \multirow{2}{*}{S4 Scale \& Proportion} & \colorbox{YellowGreen!40}{size\_ratio()} & object-to-room size ratio (\%); & \multirow{2}{*}{T2}\\
     & & \colorbox{Apricot!40}{size\_check()} & LLM judgement on absolute size plausibility \\
     \cmidrule{2-5}
     & S5 Visual Composition & \colorbox{RoyalBlue!30}{visual\_balance\_check()} & VLM judgement on top-down visual balance assessment & T5\\
     \midrule
     \multirow{9}{*}{\rotatebox[origin=c]{90}{\parbox[c]{1cm}{\centering Ergonomic}}} & \multirow{2}{*}{E1 Circulation} & \colorbox{YellowGreen!40}{pathfinding()} & the path (list of $(x,z)$ world-coordinate waypoints) or null; & \multirow{2}{*}{T2} \\
     & & \colorbox{YellowGreen!40}{path\_width()} & minimum clearance (m) and bottleneck position ($x,z$)\\
     \cmidrule{2-5}
      & \multirow{2}{*}{E2 Interaction Clearance} & \colorbox{YellowGreen!40}{articulation\_zone()} & minimum clearance within swing arc (m); & \multirow{2}{*}{T4} \\
     & & \colorbox{YellowGreen!40}{chair\_clearance()} & front and rear clearance distances from chair (m)\\
     \cmidrule{2-5}
      & E3 Reachability & \colorbox{Apricot!40}{reach\_check()} & LLM judgement on reachability given user's attributes and \colorbox{YellowGreen!40}{object\_info()}& T4 \\
      \cmidrule{2-5}
     & E4 Body Fit \& Posture & \colorbox{Apricot!40}{posture\_check()} & LLM judgement on posture given user's attributes and object's size & T4 \\
     \midrule
     \multirow{12}{*}{\rotatebox[origin=c]{90}{\parbox[c]{1cm}{\centering Activity}}} & \multirow{3}{*}{A1 Activity Zone} & \colorbox{YellowGreen!40}{free\_floor\_area()} & free area within the zone (m$^2$) ; & \multirow{3}{*}{T3} \\
     & & \colorbox{Apricot!40}{object\_in\_zone()} & LLM judgment on if related objects are correctly positioned within the zone;\\
     & & \colorbox{Apricot!40}{activity\_support\_check()} & LLM judgment on if related objects can support the required activity well\\
     \cmidrule{2-5}
     & A2 Sightlines \& Privacy & \colorbox{Apricot!40}{inbetween\_check()} & LLM judgement on if objects blocks between two points based on \colorbox{YellowGreen!40}{object\_info()}& T3 \\
      \cmidrule{2-5}
     & \multirow{2}{*}{A3 Workflow Sequencing} & \colorbox{YellowGreen!40}{total\_path\_length()} & total path length across activity sequence (m); & \multirow{2}{*}{T5} \\
     & & \colorbox{Apricot!40}{workflow\_check()} & LLM judgment on workflow order\\
     \cmidrule{2-5}
     & A4 Multi-activity Compat & \colorbox{Apricot!40}{multi\_activity\_check()} & LLM judgement on if space supports multi-activity simultaneously reusing tools for A1 & T5 \\
     \midrule
     \multirow{12}{*}{\rotatebox[origin=c]{90}{\parbox[c]{1cm}{\centering Environmental}}} & N1 Natural Light Access & \colorbox{YellowGreen!40}{window\_obs\_ratio()} & object within proximity radius blocking window ratio (\%) & T6 \\
     \cmidrule{2-5}
     & \multirow{2}{*}{N2 Glare Prevention} & \colorbox{YellowGreen!40}{screen\_window\_info()} & angle and distance between screen and window; & \multirow{2}{*}{T5} \\
     & & \colorbox{Apricot!40}{glare\_check()} & LLM judgment on glare risk\\
     \cmidrule{2-5}
     & \multirow{2}{*}{N3 Acoustic Separation} & \colorbox{YellowGreen!40}{zone\_distance()} & distance between two zones (m); & \multirow{2}{*}{T5} \\
     & & \colorbox{Apricot!40}{acoustic\_check()} & LLM judgment on acoustic risk\\
     \cmidrule{2-5}
     & \multirow{2}{*}{N4 Ventilation \& Thermal} & \colorbox{YellowGreen!40}{vent\_obs\_ratio()} & object within proximity radius blocking vent ratio; & \multirow{2}{*}{T6} \\
     & & \colorbox{YellowGreen!40}{distance\_check()} & safe distance guarantee (bool)\\
    \bottomrule
    \end{tabular}
    \label{tab:tools}
    \vspace{-1em}
\end{table*}

In practice, layout requirements are not universal. A mobility-limited user and a frequent entertainer sharing the same room type will impose fundamentally different demands on furniture placement, clearance, and zone allocation. 
We capture this variability through \textbf{activity~$\times$~persona} combinations derived from the user's functional specifications, which together determine which constraints are relevant and how their thresholds are parameterized for a given scene. 
We detail this process further in Section~\ref{sec:method}.
We ground our constraints taxonomy in established interior design literature and organize constraints into four categories: Spatial, Ergonomic, Activity, and Environmental. We illustrate these constraints in Figure~\ref{fig:constraints} and Table~\ref{tab:tools}:

\paragraph{Spatial}
Forming the foundation upon which all other constraint categories depend, these constraints govern the placement and relational arrangement of furniture within the room. 
\textbf{Geometry Validity (S1)} requires that every furniture piece fit within the room boundary without overlapping adjacent objects, except where a containment or nesting relationship is explicitly defined between them; \textbf{Boundary \& Attachment (S2)} specifies that floor-based items rest on the floor plane, wall-mounted objects attach to the correct surface, and large case goods align to the nearest wall; \textbf{Spatial Relationships (S3)} captures the grouping logic central to interior design practice, where functionally paired objects must be placed in proximity and with appropriate relative orientation; \textbf{Scale \& Proportion (S4)} ensures furniture size is commensurate with room dimensions and neighboring objects, preventing pieces from dominating or failing to define the space; \textbf{Visual Composition (S5)} captures aesthetic principles such as focal point orientation, visual balance, and alignment, ensuring the arrangement reads as intentional rather than arbitrary.

\begin{figure*}[t]
    \centering
    \includegraphics[width=\linewidth]{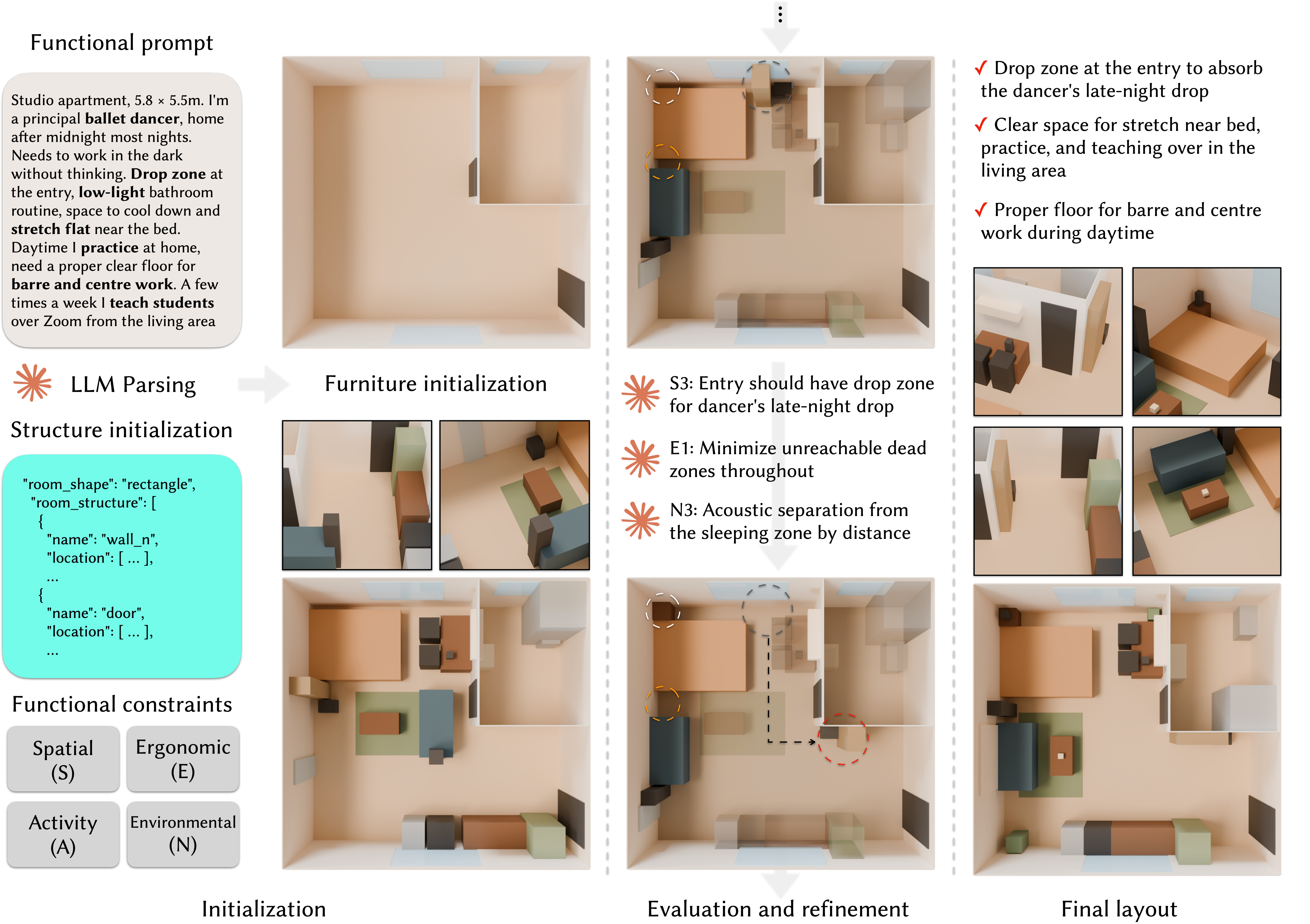}
    \caption{\textbf{Overall Pipeline:} Given a functional prompt, \textsc{Function2Scene} generates 3D indoor scene layout through iteratively evaluation and refinement based on functional constraints. }
    \label{fig:pipeline}
    \vspace{-1.5em}
\end{figure*}

\paragraph{Ergonomic}
Where spatial constraints establish what fits, ergonomic constraints ensure the space can be safely and comfortably navigated and used, with thresholds parameterized to the physical needs and abilities of the target persona. \textbf{Circulation (E1)} requires that primary pathways maintain a minimum clear width for unobstructed passage, with thresholds elevated for users relying on mobility aids; \textbf{Interaction Clearance (E2)} ensures that all articulated elements have full action zones free of obstruction, and that seating has adequate pull-out space behind it; \textbf{Reachability (E3)} constrains frequently used objects and controls to fall within the user's operational height range, accounting for seated versus standing use and any physical limitations; \textbf{Body Fit \& Posture (E4)} requires work surface heights, seat dimensions, and monitor distances to conform to anthropometric standards for sustained comfortable use.

\paragraph{Activity}
Beyond physical validity and ergonomic fit, a layout must support the specific tasks the user performs in the space, which is the primary vehicle through which the \textbf{activity~$\times$~persona} combination shapes the constraint pool. \textbf{Zone Allocation (A1)} ensures each primary activity has a dedicated zone of sufficient size, equipped with the relevant furniture and kept clearable and accessible when needed; \textbf{Multi-Activity Compatibility (A2)} demands the layout support zone transformation without full reorganization when a space serves more than one purpose; \textbf{Sightlines and Privacy (A3)} encodes directional requirements, where certain tasks demand unobstructed views toward an entry or a child's play area, while others call for visual shielding from the rest of the room; \textbf{Workflow Sequencing (A4)} arranges objects to match their order of use, avoiding backtracking and cross-circulation.
 
\paragraph{Environmental}
Furniture placement shapes not only how a room is used but how it feels, through its effect on light, sound, and thermal comfort. \textbf{Natural Light Access (N1)} asks that furniture arrangement preserve daylight reach into primary activity zones; \textbf{Glare Prevention (N2)} guards against screens and resting surfaces receiving direct window light during typical use hours; \textbf{Acoustic Separation (N3)} requires for noise-generating activities to be spatially buffered from quiet zones through distance, furniture mass, or deliberate zone boundaries; \textbf{Ventilation \& Thermal Comfort (N4)} ensures furniture does not block windows, and that temperature-sensitive activities are positioned away from cold drafts or direct heat sources.

Constraints are assigned to one of six priority tiers (T1 to T6) as indicated in Table~\ref{tab:tools}, where lower tiers must be satisfied before higher tiers are considered. Details of how each constraint is evaluated are described in Section~\ref{subsec:cons-eval}.

\section{Method}
\label{sec:method}

Our system is a function-driven interior design agent that takes a \emph{functional prompt} as input, which captures the user's living needs as input distilled through multi-turn conversations, and produces a furniture layout tailored to how they actually inhabit the space, guided by design principles derived from those needs.

As illustrate in Figure~\ref{fig:pipeline}, the pipeline proceed through two main stages: (1) the \emph{Initialization} stage, which parses the user's functional prompt into a parsed scene description and a list of functional constraints, constructs the room structure with user verification, and generates an initial furniture layout; and (2) the \emph{Constraints-based Evaluation and Refinement} stage, which iteratively evaluates the layout against constraints in priority order using specialized tools, and applies targeted adjustments guided by design principles. Following a similar prompting strategy in Holodeck~\cite{yang2024holodeck}, an LLM prompt is designed for each stage throughout the pipeline, consisting of a task description, an ouput format specification, and a one-shot example. 

In this section, we first introduce the initialization stage, which parses the user's functional prompt into an initial layout and a set of functional constraints. We then provide a detailed description of the constraints-based layout evaluation and refinement stage, which iteratively optimizes the layout toward functional and spatial coherence.

\subsection{Initialization}
The initialization stage serves as the foundation of \textsc{Function2Scene}, transforming a user-provided functional description of a scene into a structured 3D layout that serves as the starting point for the evaluation and refinement steps. It consists of three steps: parsing, room structure generation, and furniture initialization.
\paragraph{Parsing.}
Given the raw functional description of a scene, the parser extracts: (a) a structured set of constraints grounded in the 4 category constraint taxonomy defined in Section~\ref{sec:constraints}; and (b) a parsed scene description that serves as an LLM-friendly reformulation of the original input, analyzed and restructured with functional constraints considerations in mind to produce a more rational and unambiguous description in a style closer to that of current text-to-scene generation models. Together, these two outputs ensure that the original functional intent is interpreted once and propagated consistently throughout the pipeline. 

\paragraph{Room Structure Generation.}
Given the parsed description, \textsc{Function2Scene} initializes the room structure through a set of LLM prompts. The room structure is encoded in a custom JSON-based Domain-Specific Language (DSL), as shown in the cyan box in Figure~\ref{fig:pipeline}, which provides a well-defined representation of walls, floors, ceilings, doors and windows, with explicit geometric attributes such as dimension and coordinates, and semantic attributes such as orientation encoded as a facing direction. This structural output is visualized for user to verify, reflecting the natural client-designer workflow where the room structure is reviewed and approved before furnishing begins. If corrections are needed, the user can easily refine the geometry through natural language prompting, making the editing process intuitive and accessible. 

\paragraph{Furniture Initialization.}
With the verified empty room and the parsed description, \textsc{Function2Scene} directly generates an initial layout. However, as we demonstrate in our evaluation, LLM-generated layouts at this stage are fundamentally limited in their spatial reasoning: objects may overlap, violate functional adjacency requirements, or produce configurations that are physically plausible but practically unusable, as shown in the initialization stage of Figure~\ref{fig:pipeline}. The furniture initialization therefore serves as a starting point rather than a final result, motivating the constraints-based evaluation and refinement stage that follows.

\subsection{Constraints-based Evaluation and Refinement}
\label{subsec:cons-eval}
Prior works have explored constraint-based layout generation, most notably Holodeck~\cite{yang2024holodeck}, which defines a set of spatial relational constraints (e.g. \textit{in front of}, \textit{near}, \textit{face to}) and optimizes object placements to satisfy them. While effective for basic spatial arrangement, such approaches are 
restricted to geometric relations between objects, leaving ergonomic, 
activity, and environmental demands unaddressed.

\textsc{Function2Scene} extends this idea to a richer, 4-category constraint set derived directly from the user's functional input, as detailed in Section~\ref{sec:constraints}. Rather than optimizing all constraints simultaneously through a solver, we adopt an LLM-driven iterative evaluation loop that assesses constraints sequentially in priority order, as specified in Table~\ref{tab:tools}. 

\paragraph{Constraint Evaluation.}
For each constraint, specialized tools are invoked to retrieve structured spatial information, which the LLM interprets to determine whether the constraint is satisfied. Specifically, the agent first interprets the constraint description in the context of the current layout state, then selects and invokes the appropriate tool(s) such as \texttt{pathfinding()}, \texttt{visual\_balance\_check()}, or \texttt{posture\_check()} to retrieve varied forms of feedback, ranging from numeric measurements and traversal paths to natural language assessments from numeric algorithm, VLM or LLM-based tools. The LLM then interprets these results against the constraint requirements and produces a justification along with a proposed refinement step if the constraint is not met.  

\paragraph{Layout Refinement.}
For each unsatisfied constraint, the LLM generates a targeted refinement action based on the justification produced in the evaluation step. Refinement guidance is grounded in a set of design principles~\cite{panero1962anatomy} covering both universal spatial standards and room-specific human factor recommendations. Universal principles include maintaining a minimum \texttt{36''} primary circulation path and preserving door swing clearances. Room-specific standards further constrain the refinement: in the bedroom, a minimum \texttt{2'0''--3'0''} side clearance around the bed is required for access and bed-making; in the dining room, at least \texttt{3'4''} must be preserved behind seated diners for service circulation; and in the living room, conversation groups should be arranged within 8 feet of each other for comfortable interaction. These standards inform how furniture should be repositioned, reoriented, or resized. Each adjustment is applied locally to avoid disrupting already-satisfied constraints, and the affected constraint is re-evaluated before the agent proceeds to the next one. 

\paragraph{Termination.}
Constraints are evaluated sequentially in priority order. If resolving a constraints would introduce violations in higher-priority constraints that have already been satisfied, the constraint is skipped. Once all constraints have been assessed, the Tier~1 spatial constraints are re-evaluated to verify that no adjustments made in later stages have compromised foundational layout quality. The final layout is then returned as output. 
\section{Results and Evaluation}

In this section, we present qualitative and quantitative results demonstrating the ability of our method in generating visually and functionally plausible indoor scenes given functionality-focused natural language prompts. We compare against three representative LLM-based layout generation methods: Holodeck~\cite{yang2024holodeck}, iDesign~\cite{ccelen2024idesign}, and LayoutVLM~\cite{sun2025layoutvlm}, which cover the current landscape of language-driven indoor scene synthesis from persona-aware generation to open-ended spatial instruction following. Our evaluation focuses on how well each approach captures persona-specific functional requirements and lifestyle-driven spatial organization that standard benchmarks often overlook.

\subsection{Data}
We curated 30 real interior design cases from \emph{Architectural Digest}~\cite{architecturaldigest2026}, an internationally recognized magazine and authority on interior design. Each case describes a distinct room designed around a specific occupant persona, resulting in a diverse dataset spanning 10 room types, including bedrooms, kitchens, living rooms, dining rooms, studios/ateliers, a home library, a guestroom, a nursery, a great room, and a mezzanine; 30 unique personas ranging from a retired couple and a chef to a drag queen, a child with autism, and a YouTuber. This breadth ensures coverage of varied functional needs, aesthetic preferences, lifestyle contexts, and demographic profiles. 

\subsection{Perceptual Study}
To evaluate how well different layout generation methods satisfy the functional requirements specified in user scene prompts, we conducted a two-alternative forced-choice (2AFC) perceptual study. We recruited 30 participants through Prolific~\cite{prolific2026}, with diverse backgrounds prior experience with AI evaluation tasks. 
Each participant completed 30 scene comparisons, where each comparison presented a room brief alongside rendered images of two layouts in randomized order, and selected the more functional layout for the described occupants.
We also randomly insert 5 attention checks that compares against randomly generated, implausible layouts, and filter out participants who fail any of these checks.

We have a total of 10 comparison conditions (6 baselines, 4 ablations), resulting in 3 answers per scene and condition.  
For each comparison, we aggregate responses across all valid participants who saw that pair and report the proportion of selections in favour of our method. 
To reduce evaluation time and focus judgments on layout quality rather than scene-level details, participants were instructed to prioritize structural validity, such as furniture blocking doorways or objects extending outside room boundaries before considering brief-specific criteria.

\begin{figure*}
    \centering
    \includegraphics[width=0.99\linewidth]{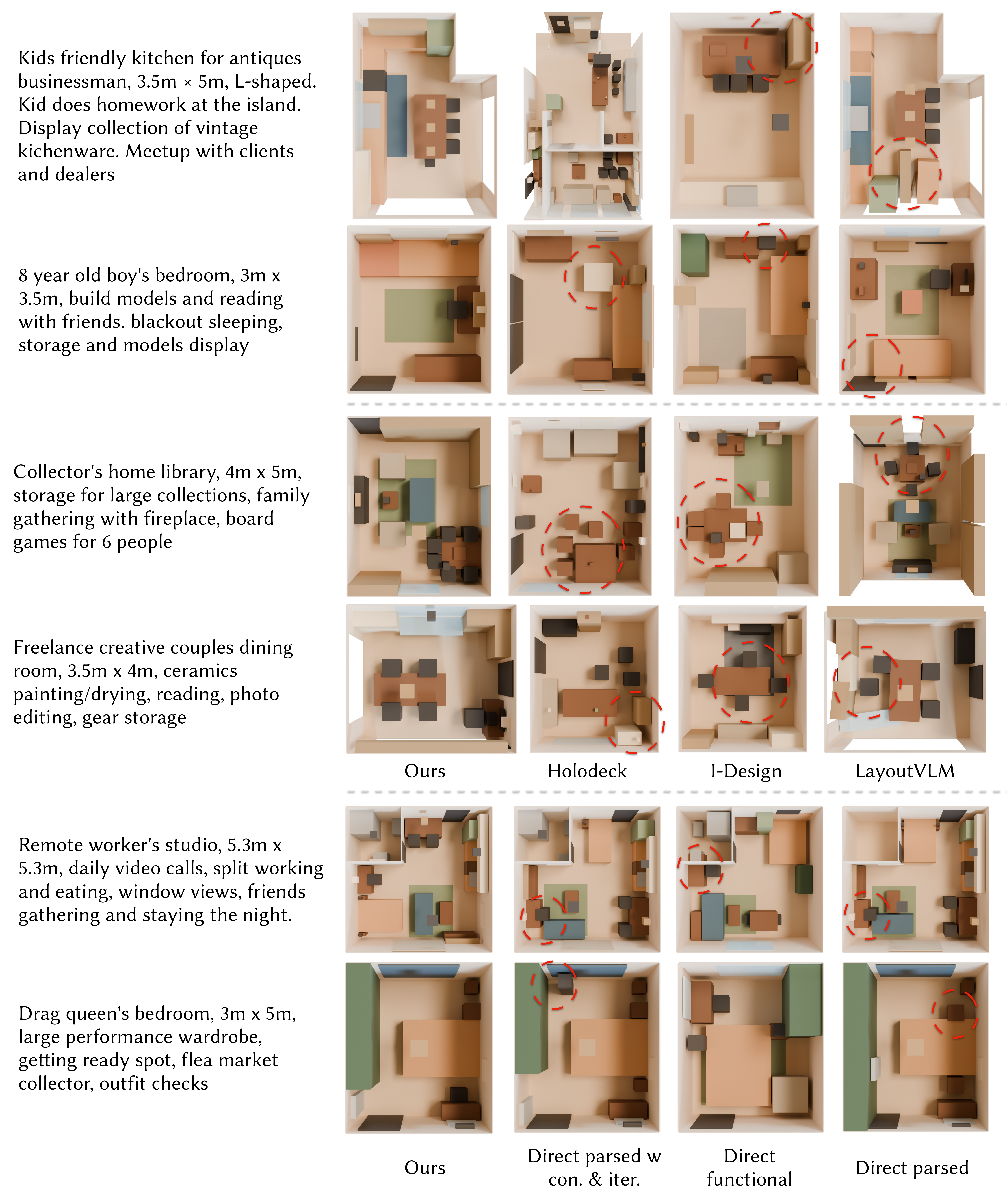}
    % \vspace{-7em}
    \caption{Qualitative comparisons of our method against various comparison conditions. Top two rows: baselines with original functional prompts; middle two rows: baselines with our parsed specifications; bottom two rows: ablations, from left to right: w/ parsed input and iterative refinement, with original prompt and no iterative refinement, with parsed input and no iterative refinement.}
    \label{fig:fig_only_1}
\end{figure*}

\begin{figure*}
    \centering
    \includegraphics[width=0.99\linewidth]{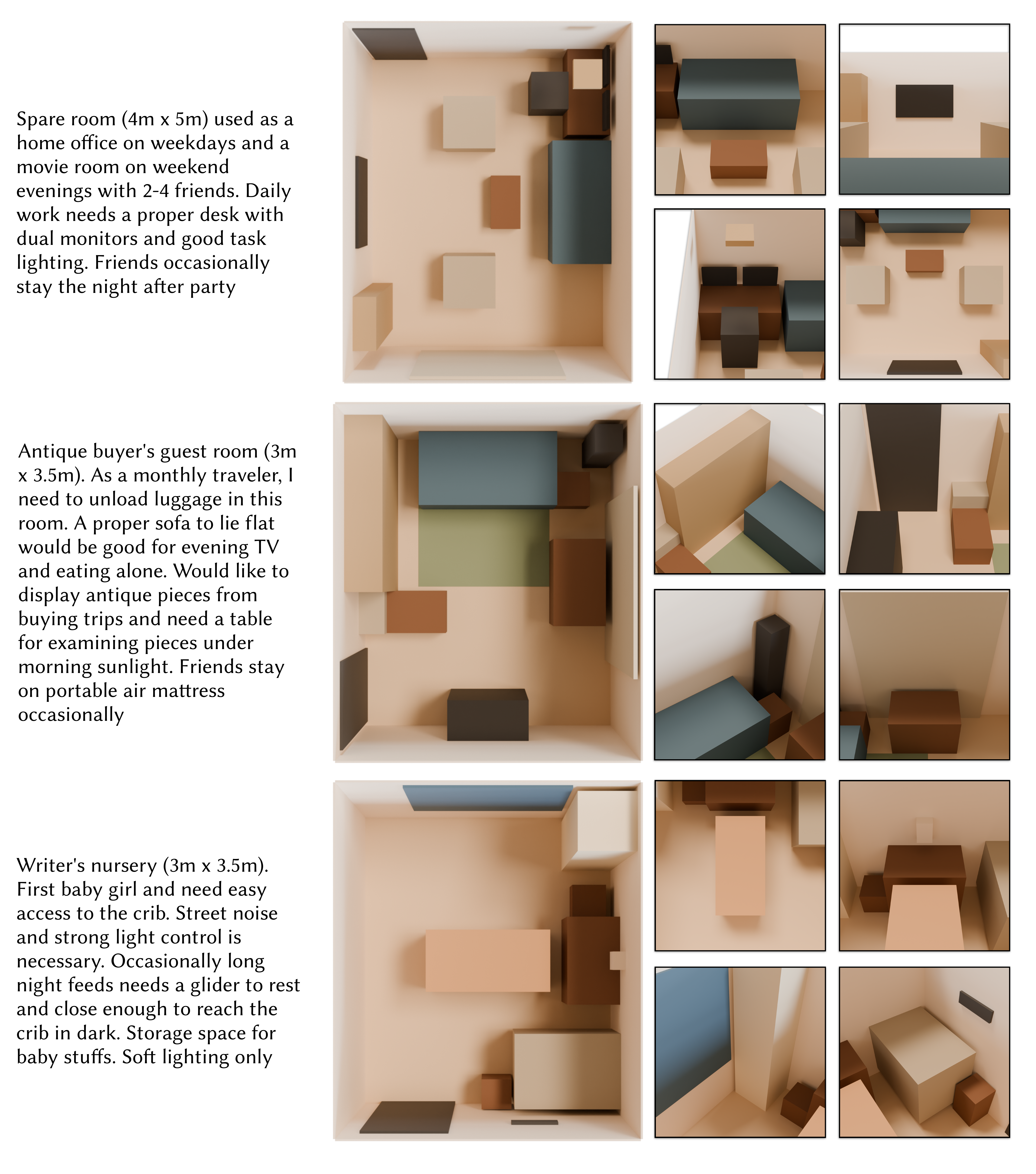}
    % \vspace{-7em}
    \caption{Functional scenes generated by our method, along with zoomed in highlights. The input prompts are truncated due to space constraints. Please refer to the supplementary materials for all qualitative results, along with visualization of all intermediary optimization steps.}
    \label{fig:fig_only_2}
\end{figure*}

Table~\ref{tab:user_study} shows the results of this experiment, and Figure~\ref{fig:fig_only_1} shows some qualitative comparisons between generated layouts. Overall, participants preferred layouts generated using our method across all baselines and prompt conditions, with an aggregate preference rate of 94.3\%. Against Holodeck, our method was preferred in 92.2\% and 88.9\% of trials under functional and parsed prompts respectively. Against iDesign, preference rates reached 94.4\% and 98.9\%, with the parsed condition yielding the highest score across all comparisons. Against LayoutVLM, our method was preferred in 96.7\% and 94.4\% of trials. Figure~\ref{fig:teaser} and Figure~\ref{fig:fig_only_2} provides a closer look at our generated layouts, highlighting fine-grained constraint satisfaction across spatial and functional requirements.
Please refer to the supplementary material for more detailed visualization of our results and comparisons.

Furthermore, Table~\ref{tab:ablation} presents ablation study results examining the contribution of each component in our pipeline. Notably, retaining iterative updates without evaluation tools performs worst than removing both, indicating that iterative refinement is counterproductive without grounded spatial feedback to guide it. Additionally, prompt format has negligible effect when tools are absent, confirming that richer constraint representations only enforces their benefit when paired with the tool set that measures them. These results establish evaluation tools as the critical enabler of our pipeline.

%Details of the study setup are presented in the supplement. 
\begin{table}[t]
  \centering
  \footnotesize
  \caption{2AFC study results comparison our method with baselines.
           Each baseline is ran with both the original functional prompt
           and our parsed scene description and constraints.} 
  % \vspace{-1em}
  \label{tab:user_study}
  \setlength{\tabcolsep}{7pt}
  \begin{tabular}{llc}
    \toprule
    Method & Prompt & \% Ours preferred \\
    \midrule
    \multirow{2}{*}{Holodeck~\cite{yang2024holodeck}}
      & Functional   & 92.2 \\
      & Parsed  & 88.9 \\
      \midrule
    \multirow{2}{*}{iDesign~\cite{ccelen2024idesign}}
      & Functional  & 94.4 \\
      & Parsed  & 98.9 \\
      \midrule
    \multirow{2}{*}{LayoutVLM~\cite{sun2025layoutvlm}}
      & Functional  & 96.7 \\
      & Parsed & 94.4 \\
    \midrule
    Overall & --- & 94.3 \\
    \bottomrule
  \end{tabular}
\end{table}
\begin{table}[t]
  \centering
  \footnotesize
  \caption{2AFC study results comparing against ablations that uses different input and generation strategy.}
  % \vspace{-1em}
  \label{tab:ablation}
  \setlength{\tabcolsep}{7pt}
  \begin{tabular}{ccccc}
    \toprule
    Prompt format & Iterative update & Evaluation Tools & \%Ours preferred \\
    \midrule
    Functional & No & No & 83.3 \\
    Parsed & No & No & 83.3 \\
    Functional & Yes & No & 78.9 \\
    Parsed & Yes & No & 80.0\\
    \midrule
    Parsed & Yes & Yes & \emph{Ours}\\
    \bottomrule
  \end{tabular}
  % \vspace{-2em}
\end{table}

%\begin{table}[t]
%  \centering
%  \footnotesize
%  \caption{Ablation study (2AFC). \checkmark\ indicates the component
%           is included. Rows not using \emph{parsed prompt} use \emph{functional prompt}. 
%           Random layout comparisons are excluded.}
%  \label{tab:ablation}
%  \setlength{\tabcolsep}{7pt}
%  \begin{tabular}{cccccc}
%    \toprule
%    Parsed prompt & Constraints & Tool & Iter. & Ours preferred & Win rate\\
%    \midrule
%    \checkmark & \checkmark & \checkmark & \checkmark & \multicolumn{2}{c}{\emph{Ours (full method)}} \\
%           &  &  &  \checkmark & 71 / 90  & 78.9\% \\
%    \checkmark & \checkmark &  & \checkmark  & 72 / 90  & 80.0\%\\
%    \checkmark & \checkmark &    &  & 75 / 90  & 83.3\% \\
%           &        &        &  & 75 / 90  & 83.3\% \\
%    \midrule
%    \multicolumn{3}{l}{Overall} &  & 293 / 360 & 81.4\%\\
%    \bottomrule
%  \end{tabular}
%\end{table}
\section{Conclusion}
\label{sec:conslusion}
In this paper, we presented Function2Scene, a framework for generating indoor layouts from functional specifications.
By focusing on functionality, we take a first step towards designing a LLM-driven scene generation pipeline that better suits real interior design workflows, demonstrating that a well-designed taxonomy of functional design principles, combined with a LLM-driven iterative pipeline, can produce higher quality, more functional scenes than those generated by prior works.

There exist many opportunities in further extending our method:
as stated in the introduction, our method starts from a professionally written, detailed functional specification.
Real design workflows, however, usually begin with user demands that are much vaguer and shorter: users need designers' help to discover, articulate, and refine their needs through multiple rounds of conversation and feedback.
A conversational interface that helps non-expert users arrive at a detailed functional specification would complete the upstream half of the design workflow that feeds into our method;
while our constraint taxonomy provides a general framework for evaluating functionality, our current verification protocol heavily relies on basic numeric checks and LLM queries. 
They could be made much more powerful with more constraint-specific tools, for example, embodied simulation with articulated models, physically accurate lighting and acoustic estimations, or a domain-specific language for expressing distance and dimension requirements in more semantic manners;
more broadly, our framework currently operates within a fixed architectural shell in residential environments, co-optimizing room shape, openings, and partitions alongside furniture placement would better capture the full scope of interior design.

\clearpage
\bibliographystyle{ACM-Reference-Format}
\bibliography{ref}

\clearpage

\clearpage
\clearpage
\appendix
We provide a GUI showing the completed iterations and results in \texttt{index\_static.html} in the supplementary. We also provide a GUI showing all qualitative comparisons in \texttt{qualitative.html}. 
\section{Layout Representation}
To formally specify room layouts, we define a \textbf{Domain-Specific Language
(DSL)} encoded in JSON. This DSL provides a structured, machine-readable representation of a room's architectural surfaces and furnishings, designed to be both human-authored and amenable to automated layout validation and 3D rendering.

The DSL represents a room as a JSON object with two top-level arrays: \texttt{room\_structure} and \texttt{furniture}. The \texttt{room\_structure} array enumerates all architectural surfaces: walls, floor, ceiling, door, and window, while \texttt{furniture} enumerates all
movable objects placed within the room. A short \texttt{\_convention} block at the top of each file defines the shared coordinate system and field semantics that apply throughout.

All geometry uses a right-handed coordinate system with axes $+X$ (east), $-X$ (west), $+Y$ (up), and $+Z$ (south). Every surface and object carries a \texttt{location} field giving its centroid in world space, and a \texttt{dimensions} field expressed as \texttt{[width, height, depth]} in metres in the element's local frame. Orientation is encoded by a \texttt{facing} field---an integer bearing in $[0^\circ, 360^\circ)$ representing the direction the element's outward or inward normal points, following compass convention ($0^\circ = \text{north}$, $90^\circ = \text{east}$, $180^\circ = \text{south}$, $270^\circ = \text{west}$). Walls use their inward normal as the facing value; floor and ceiling omit the field entirely, as their normals are fixed by convention. Openings such as doors and windows are represented either as standalone entries with physical dimensions, or as \texttt{holes} sub-arrays on the parent wall, giving a 2D offset and size within that wall's local surface frame.

Each furniture entry additionally carries an \texttt{orientation} tag: \texttt{"directional"} for objects with a meaningful front face, \texttt{"axial"} for objects symmetric about one axis (e.g.\ a rug), and \texttt{"symmetric"} for fully rotationally symmetric objects such as a ceiling light. This distinction allows downstream rendering and constraint-solving code to apply appropriate symmetry assumptions when validating placement rules. The top-level \texttt{description} field is present only in refined or final scenes, where it records the set of layout constraints that have been resolved in the current iteration. It serves as a human-readable invariant summarising the design intent that the scene satisfies: for example, asserting that the footprint of every personal storage unit is large enough to accommodate one occupant's books, devices, and weekend gear without overflow.

\begin{figure}[ht]
\centering
\caption{Sample layout DSL file.}
\begin{lstlisting}[language=json, basicstyle=\ttfamily\small,
  breaklines=true, frame=single]
{
  "description": "S3 RULE: ...",
  "room_structure": [
    {
      "name": "wall_n",
      "location": [0, 1.2, -2.26],
      "dimensions": [3.5, 2.4, 0.02],
      "facing": 180,
      "color": "#F2F0EB",
      "holes": [{"location": [1.1, -0.15], "dimensions": [0.9, 2.1]}]
    },
    ...
  ],
  "furniture": [
    {
      "name": "bunk_bed",
      "orientation": "directional",
      "location": [0.75, 0.825, -1.75],
      "dimensions": [2.0, 1.65, 1.0],
      "facing": 180,
      "color": "#C49464"
    },
    ...
  ]
}
\end{lstlisting}
\end{figure}

\section{Perceptual Study Details}
\paragraph{Study Design.}
We conducted a two-alternative forced-choice (2AFC) perceptual study to evaluate layout quality across 30 room scenes. Each scene was associated with one output from our method and outputs from 10 baseline methods or ablations (10 comparison types in total). To ensure each participant evaluated a diverse cross-section of scenes, each survey instance contained exactly 30 pairs, with each pair drawn from a different scene. Across all 30 pairs, the comparison method for each pair was sampled from the pool of 10 comparison types, ensuring broad coverage.

\paragraph{Attention Checks.}
5 of the 30 pairs in each survey were attention check items, in which our method's output was compared against a randomly generated layout. These checks were distributed among the 30 pairs and served to identify inattentive respondents. Any participant who failed to correctly identify the clearly superior layout on any attention check question was excluded from analysis.

\begin{figure*}[h]
    \centering
    \includegraphics[width=0.9\linewidth]{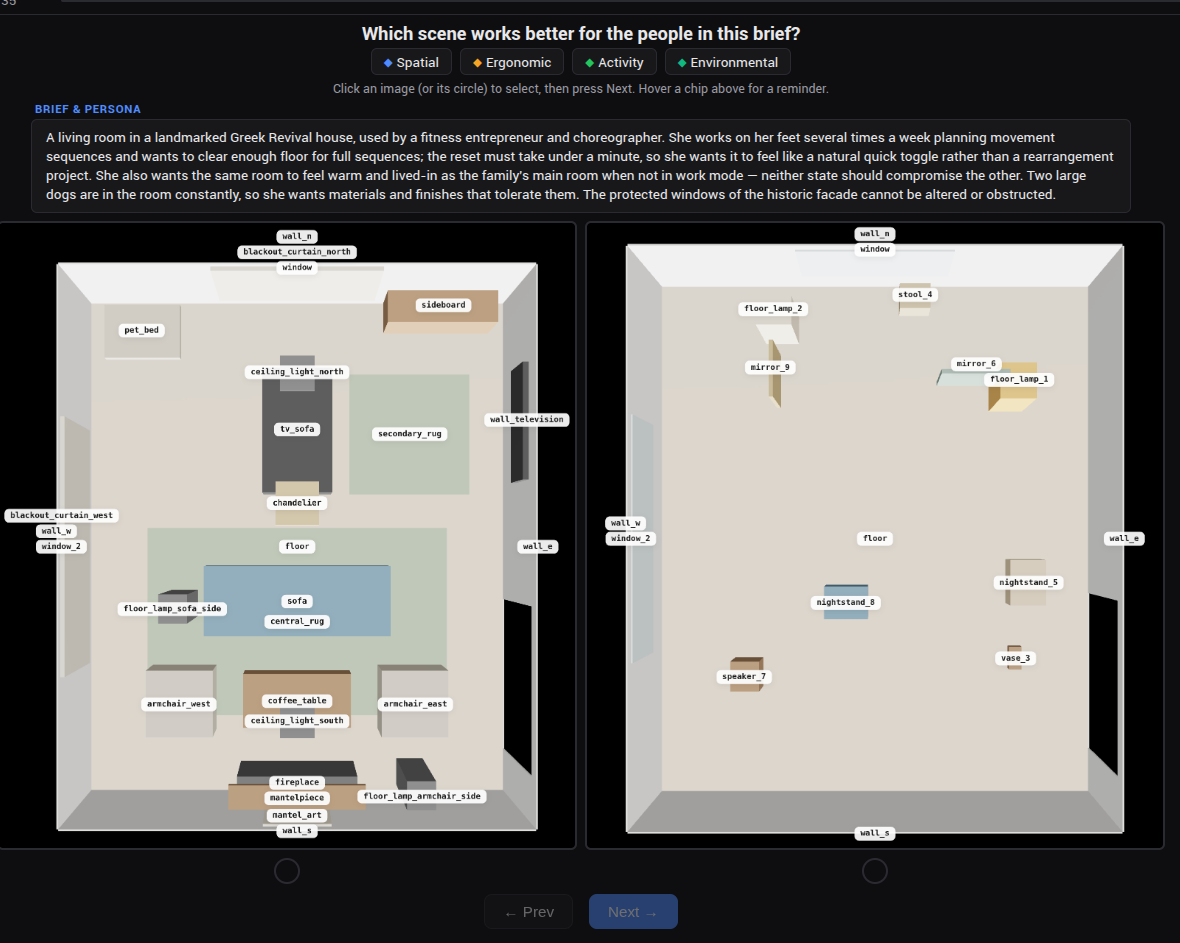}
    \vspace{-1em}
    \caption{Perceptual study interface.}
    \label{fig:supp-user-2}
\end{figure*}

\paragraph{Survey Variants.}
We fixed 10 distinct survey variants, each containing a different assignment of comparison methods across the 30 scenes. Each variant was completed by 3 participants, yielding a target of 30 submissions. Each survey was presented as a web-based interface (screenshots provided in Figure~\ref{fig:supp-user-2}) in which participants viewed a brief describing the room's intended occupants and use cases alongside two rendered scene images, and selected the better layout.

\paragraph{Recruitment.}
Participants were recruited via Prolific under an AI task evaluation study category. Eligibility criteria required: normal or corrected-to-normal vision, no colour blindness, current country of residence in the United States or Canada, at least 100 prior Prolific submissions, and at least 10 prior AI evaluation task submissions. These criteria were chosen to ensure participants had prior experience with structured evaluation tasks and could reliably perceive visual differences in the rendered scenes.

\paragraph{Participants.}
We collected 45 submissions in total, of which 32 passed all attention checks and were retained for analysis (valid response rate: 71.1\%). The 32 valid participants varied in age, racial background, and occupation, providing a demographically diverse sample. We remove 2 repeated results. 

\begin{figure}[t]
    \centering
    \includegraphics[width=0.8\linewidth]{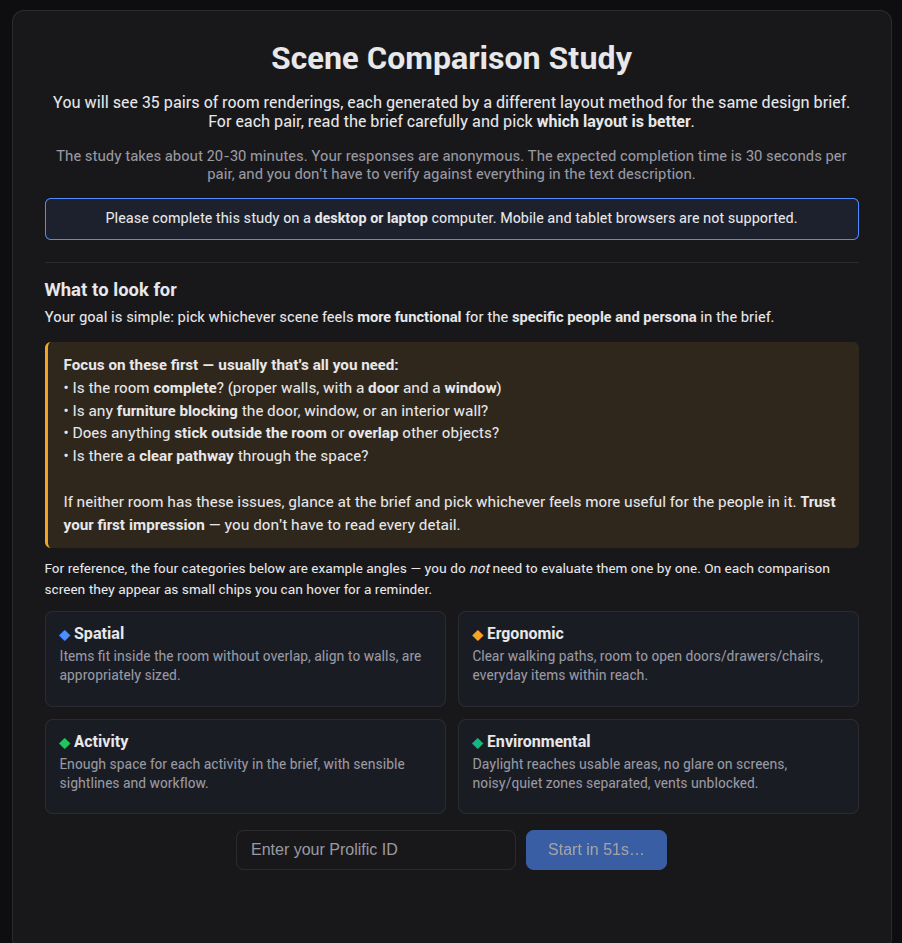}
    \vspace{-1em}
    \caption{Perceptual study introduction.}
    \label{fig:supp-user}
\end{figure}

\paragraph{Interface.} 
The study interface presented each pair of rendered scenes side by side, accompanied by the room brief and persona description. Participants selected their preferred layout by clicking the image or a radio button below it, then advanced to the next pair. Four evaluation dimensions: spatial validity, ergonomics, activity support, and environmental quality were described in the onboarding instructions and available as hover-over reminder chips during the study as shown in Figure~\ref{fig:supp-user}. Participants were instructed to prioritize structural issues (missing walls, doors, or windows; furniture blocking egress; objects extending outside room boundaries; lack of clear pathways) before consulting brief-specific criteria. The estimated completion time was 20–30 minutes (approximately 30 seconds per pair).

\section{Implementation Details}
\subsection{Initialization}
The room initialization pipeline converts a natural-language room description into a structured JSON scene through a sequence of discrete, LLM-driven steps. Each step is implemented as a structured prompt that instructs the language model to perform a narrowly scoped transformation, and intermediate outputs are saved to disk so that a human reviewer can inspect and correct the structure before the next stage begins.
\paragraph{Room shell generation.}
The pipeline opens by classifying the room as either rectangular or L-shaped based on linguistic cues in the description. Phrases indicating a structural recess: such as "alcove," "nook," "missing corner," or explicit references to two rectangular zones meeting at right angles, trigger the L-shaped path; all other descriptions default to a rectangle. Once the shape is determined, the shell prompt generates the room's structural envelope: four walls for a rectangle, or six walls tracing the L's perimeter for the L-shaped variant. Both formats share a common coordinate system (origin at the floor center, +X east, +Z south, +Y up) and produce a consistent JSON schema with named wall entries, a floor polygon or slab, and a ceiling. Critically, every \texttt{holes} array is left empty at this stage, and no openings of any kind are cut into the walls.
\paragraph{Door and window placement.}
With the shell confirmed, the second step populates openings. Doors and windows are appended as new entries in the \texttt{room\_structure} array, each carrying world-coordinate position, dimensions, facing direction, and color. Simultaneously, a matching hole record is written into the parent wall's \texttt{holes} array using the wall's local coordinate frame, accounting for the fact that the local horizontal axis maps to different world directions depending on the wall's facing. The step enforces a set of placement constraints: openings must fit within the wall's extents, must not overlap one another, and windows must respect a mandatory 0.2 m header clearance below the wall top. Standard size presets are provided for both doors (single 0.9 m, double 1.5 m) and windows (standard through panoramic), with the system defaulting to larger windows to avoid a cramped feel.
\paragraph{Human verification.}
The pipeline deliberately pauses after door and window placement. The shell and the populated structure are written as separate files so that a reviewer can compare them and make manual edits: adding interior partitions, repositioning openings, or adjusting dimensions - before saving the final structural baseline. This checkpoint prevents furniture placement from proceeding against a structurally incorrect room. 

A final prompt instructs the model to read the human-reviewed structural file and generate a \texttt{furniture} array describing every object in the scene. Each object entry is produced by the model with a center position, bounding-box dimensions, facing direction, and an orientation class (\texttt{directional}, \texttt{axial}, or \texttt{symmetric}) that signals how the object's facing should be interpreted by downstream renderers. For multi-functional spaces such as studio apartments, the prompt provides the model with zone checklists: sleeping, living, kitchen, bathroom, work, and dining, ensuring that every implied zone is populated with its essential objects even when the description names a zone without enumerating individual items. As with all prior steps, the model is instructed to leave the structural elements entirely untouched; furniture placement is strictly additive.

\subsection{Functional Constraints}
We define the parser to generate parsed scene description and a list of functional constraints based on the functional prompt. Here is an example of the complete constraints, more examples can be referred in the website we provided. 
\begin{lstlisting}[language=markdown, frame=single]
ZONE MAP
========
Zone                       | HARD or SOFT | Boundary defined by
Desk work area             | SOFT         | Wherever the desk, office chair, and task lamp cluster lands
Movie viewing and lounge   | SOFT         | Wherever the screen, sofa, occasional seating, and coffee table cluster lands
Guest sleeping area        | SOFT         | Wherever the sofa bed or fold-out bed is deployed; overlaps the lounge cluster when a sofa bed is used


CONSTRAINTS
===========

TIER 1 - HARD
--------------

GROUP: Outer Boundary and Interior Walls
Priority: S1 always first
-------------------------------------

S1  RULE: no floor-placed object extends beyond the outer room boundary
    APPLY TO: every floor-placed object in the scene JSON
    PASS IF: overlap_ratio = 0 for every object
    IF FAIL: translate the failing object inward along the axis of overlap until overlap_ratio = 0; recheck S1 overlap after move
    EXCEPTIONS: NONE

S1  RULE: no wall-mounted object extends beyond the outer room boundary
    APPLY TO: every wall-mounted object in the scene JSON
    PASS IF: overlap_ratio = 0 for every object
    IF FAIL: translate the failing object inward along the wall until overlap_ratio = 0; verify the object still lies flush against the wall after move
    EXCEPTIONS: NONE

S1  RULE: no two floor-placed objects overlap each other
    APPLY TO: every unordered pair of floor-placed objects in the scene JSON
    PASS IF: overlap_ratio = 0 for every pair
    IF FAIL: translate the lower-priority object in the pair away from the higher-priority object along the axis of greatest overlap until overlap_ratio = 0; recheck boundary rule after move
    EXCEPTIONS: any chair or stool and its directly paired table or desk surface may have a non-zero overlap ratio consistent with the seat tucked fully under the table apron or desk knee space; the executor determines which chairs are paired with which tables from the scene JSON cluster assignments; the ratio for every non-paired object pair must be 0


GROUP: Doors, Fixtures, and Windows
Priority: S2 > E1
-------------------------------------

--- FLOOR CONTACT AND WALL ATTACHMENT ---

S2  RULE: every floor-based object rests flat on the floor plane
    APPLY TO: every floor-placed object in the scene JSON
    PASS IF: Z = 0 for every object
    IF FAIL: snap the failing object to Z = 0
    EXCEPTIONS: NONE

S2  RULE: every wall-mounted object is flush against its backing wall
    APPLY TO: every wall-mounted object in the scene JSON
    PASS IF: angle to nearest wall <= 5 degrees for every object
    IF FAIL: rotate and translate the failing object until it lies flat against the wall surface
    EXCEPTIONS: NONE

--- WALL-MOUNT VALIDITY ---

S2  RULE: every wall-mounted object sits entirely on solid wall, not spanning a window opening, door opening, or corner void
    APPLY TO: every wall-mounted object in the scene JSON; wall_segment is the specific wall section behind the object, with window and door openings masked as non-solid
    PASS IF: the object's full horizontal and vertical span falls within solid wall only; overlap with any opening = 0
    IF FAIL: slide the object laterally along the wall until its full span clears every opening; if no solid span is long enough, flag for replacement with a narrower unit or relocation to an adjacent wall
    EXCEPTIONS: curtain rails are exempt - they are by design wider than the window and span the opening; no other wall-mounted object is exempt

--- HINGED DOORS ---

S2  RULE: the swing arc of every hinged door is clear of all objects
    APPLY TO: every door in the scene that has a physical door leaf; if door type is not stated, treat as hinged by default
    PASS IF: swing arc polygon intersects no object bounding box
    IF FAIL: move the blocking object out of the swing arc zone; if the object is fixed, flag the door as requiring a sliding conversion
    EXCEPTIONS: NONE

S2  RULE: the approach path to every hinged or sliding door is unobstructed on both sides
    APPLY TO: every hinged or sliding door in the scene JSON; entry_point is one body-width directly in front of the door on each side
    PASS IF: a clear straight path exists from entry_point to door face on both sides
    IF FAIL: move the blocking object away from the door face to restore the clear approach on the blocked side
    EXCEPTIONS: NONE

--- OPEN THRESHOLDS ---

S2  RULE: no object occupies the clear-path zone of an open threshold between two zones
    APPLY TO: every opening between zones that has no physical door leaf (open-plan archways, zone transitions described as open passages, wide openings between sub-areas, etc.)
    PASS IF: a continuous clear path exists through the full width of the opening; no object centroid falls within the opening span; clear width along the path is wide enough for comfortable passage by a standard adult
    IF FAIL: move the encroaching object laterally out of the threshold span; if the clear width is too narrow, move flanking objects outward on both sides until width is sufficient
    EXCEPTIONS: a floor lamp or small planter placed at the very edge of the threshold is acceptable if it does not reduce the clear span below comfortable passage width for a standard adult

--- WINDOW OBSTRUCTION ---

S2  RULE: no object blocks a window opening
    APPLY TO: every non-curtain, non-curtain-rail object in the scene JSON; window_opening is the full rectangular face of each window in the room
    PASS IF: overlap_ratio = 0 between every object and every window opening
    IF FAIL: move the blocking object away from the window; if the object is a low unit designed to sit below a sill, run a VLM check to confirm it does not reach the glazing - if it does, replace with a shorter unit or move it to another wall
    EXCEPTIONS: curtain rails and curtains are exempt; a radiator or built-in unit below the sill line may be reviewed by VLM to confirm it does not obstruct glazing - flag for manual check if present

--- FIXTURE ORIENTATION ---

S2  RULE: every fixture with a defined use-face is oriented so the use-face is accessible from open floor
    APPLY TO: every fixture with a defined use-face: sofa (seat faces the screen wall), occasional seating (seat faces the screen or the sofa front), desk (work surface faces a wall or window so the user has back support from open floor), television or projection screen (screen face is visible from the sofa and occasional seating)
    PASS IF: angle between use-face normal and nearest unobstructed open-floor centroid <= 45 degrees
    IF FAIL: rotate the fixture until its use-face points toward open floor; recheck S1 overlap and S2 wall-attachment after rotation
    EXCEPTIONS: NONE


TIER 2 - HIGH: Layout Plausibility
------------------------------------

GROUP: Circulation
Priority: E1 > S3
-------------------------------------

E1  RULE: every main circulation path through the room is passable for this user
    APPLY TO: every pair of occupied zones between which a resident must travel in daily use; select pairs by identifying which zones are used in sequence for each activity in the scene (entry door to desk; entry door to sofa and screen; sofa to screen; sofa to entry door for guest movement on movie nights)
    PASS IF: a clear path exists between every such pair; clear width at every point along the path is wide enough for a standard able-bodied adult and is wide enough to allow two to four guests to enter and reach the sofa cluster without single-file shuffling
    IF FAIL: move the object(s) narrowing the path outward from the path centerline until clear width is sufficient along the full length
    EXCEPTIONS: NONE

E1  RULE: the entry zone immediately inside the main door has a clear sightline into the room
    APPLY TO: the main entry door; eye position is one step inside the door at standing height for a standard adult
    PASS IF: no obstruction on the ray from entry eye to room interior centroid
    IF FAIL: move the obstructing object out of the entry sightline cone
    EXCEPTIONS: NONE

E1  RULE: both long sides of any deployed sleeping surface are reachable for the guest who sleeps there
    APPLY TO: every sleeping surface in the scene JSON (sofa bed in its unfolded state and any separate fold-out bed); check each long side independently in the deployed configuration
    PASS IF: clear space along at least one long side is wide enough for a standard adult to sit on the edge and stand up; the foot end is also reachable on foot for bedding changes
    IF FAIL: move the sleeping surface or adjacent furniture in the deployed configuration to open the blocked side
    EXCEPTIONS: a sofa bed used by a single guest may have its back long side against a wall; the front long side must remain clear in the deployed configuration

E1  RULE: dead zones behind doors and in corners do not exceed an acceptable size for a standard adult
    APPLY TO: every corner and behind-door area in the room; identify dead zones as areas bounded by walls and furniture with no access path
    PASS IF: dead zone area is small enough to be non-functional storage at most
    IF FAIL: move the furniture creating the dead zone outward from the corner to reduce dead zone size; or assign the dead zone as intentional low-access storage and flag it in the scene
    EXCEPTIONS: NONE


GROUP: Desk Work Cluster
Priority: S3 > A3
-------------------------------------

S3  RULE: objects that belong to the desk work cluster are placed in proximity to each other
    APPLY TO: the desk work cluster consisting of the desk, the two monitors that sit on the desk, the office chair, the task lamp, and any nearby office storage
    PASS IF: every object in the cluster is within expected functional proximity of its cluster partners; no object from a different cluster intrudes between cluster members
    IF FAIL: move the outlying object toward its cluster centroid until the cluster is spatially coherent
    EXCEPTIONS: NONE

S3  RULE: objects within the desk work cluster share a common axis or facing alignment
    APPLY TO: every pair within the desk work cluster that has a defined facing relationship (office chair faces desk; monitors face the user position at the desk; desk back is parallel to its supporting wall)
    PASS IF: relative orientation delta between each pair is within an acceptable facing range (parallel or perpendicular as appropriate for the object type)
    IF FAIL: rotate the misaligned object until alignment is confirmed; recheck S1 overlap after rotation
    EXCEPTIONS: NONE

S3  RULE: residual space created when the desk cluster is positioned is assessed for usability
    APPLY TO: every gap or leftover area created between the desk cluster and the walls or adjacent clusters
    PASS IF: residual area is either large enough for a meaningful function (circulation path, office storage, additional seat) or small enough to be negligible; no awkward mid-room residual strip exists
    IF FAIL: shift the cluster to reduce awkward residual strips; or assign the residual area to a named function and flag it for potential use
    EXCEPTIONS: NONE

S3  RULE: the footprint or span of every office storage unit is large enough to hold the office supplies designated for the desk zone without overflow
    APPLY TO: every office storage object near the desk cluster (drawer unit, shelf, filing cabinet, etc.)
    PASS IF: storage unit capacity is sufficient for the designated load based on footprint and volume ratio
    IF FAIL: replace the storage unit with a larger unit, or add an additional unit of the same type in adjacent wall space
    EXCEPTIONS: NONE

S3  RULE: adjacent wall space near the desk cluster is evaluated for expansion capacity if primary office storage proves insufficient
    APPLY TO: the desk cluster; residual_wall_span is the wall length adjacent to the cluster not yet occupied by furniture
    PASS IF: residual wall span is noted and its capacity for an additional shelf, cabinet, or rack unit is flagged
    IF FAIL: flag the desk storage zone as expansion-constrained; recommend alternative storage locations in the scene
    EXCEPTIONS: NONE


GROUP: Movie Viewing and Lounge Cluster
Priority: S3 > A3
-------------------------------------

S3  RULE: objects that belong to the movie viewing and lounge cluster are placed in proximity to each other
    APPLY TO: the movie viewing and lounge cluster consisting of the television or projection screen, the sofa or sofa bed, any occasional seating (armchairs, floor cushions), the coffee table or low side surface, and the rug under the cluster
    PASS IF: every object in the cluster is within expected functional proximity of its cluster partners; no object from a different cluster intrudes between cluster members
    IF FAIL: move the outlying object toward its cluster centroid until the cluster is spatially coherent
    EXCEPTIONS: NONE

S3  RULE: objects within the movie viewing and lounge cluster share a common axis or facing alignment
    APPLY TO: every pair within the lounge cluster that has a defined facing relationship (sofa faces screen; occasional seating faces screen or sofa front; coffee table aligns with sofa front; rug aligns with sofa-screen axis)
    PASS IF: relative orientation delta between each pair is within an acceptable facing range (parallel or perpendicular as appropriate for the object type)
    IF FAIL: rotate the misaligned object until alignment is confirmed; recheck S1 overlap after rotation
    EXCEPTIONS: NONE

S3  RULE: residual space created when the lounge cluster is positioned is assessed for usability
    APPLY TO: every gap or leftover area created between the lounge cluster and the walls or the desk cluster
    PASS IF: residual area is either large enough for a meaningful function (circulation between desk and sofa, additional occasional seating, storage) or small enough to be negligible; no awkward mid-room residual strip exists
    IF FAIL: shift the cluster to reduce awkward residual strips; or assign the residual area to a named function and flag it for potential use
    EXCEPTIONS: NONE

S3  RULE: the footprint or span of every storage unit near the lounge cluster is large enough to hold the items designated for the lounge and guest sleep zone without overflow
    APPLY TO: every storage object placed near the lounge cluster intended to hold bedding for overnight guests, throws, remotes, and movie-night supplies
    PASS IF: storage unit capacity is sufficient for the designated load based on footprint and volume ratio
    IF FAIL: replace the storage unit with a larger unit, or add an additional unit of the same type in adjacent wall space
    EXCEPTIONS: NONE

S3  RULE: adjacent wall space near the lounge cluster is evaluated for expansion capacity if primary lounge and bedding storage proves insufficient
    APPLY TO: the lounge cluster; residual_wall_span is the wall length adjacent to the cluster not yet occupied by furniture or by the screen wall
    PASS IF: residual wall span is noted and its capacity for an additional shelf, cabinet, or sideboard is flagged
    IF FAIL: flag the lounge storage zone as expansion-constrained; recommend alternative storage locations in the scene
    EXCEPTIONS: NONE


STANDALONE
-------------------------------------

S4  RULE: every piece of furniture has a footprint proportional to the room area and to the other objects around it
    APPLY TO: every floor-placed object; compare each object's footprint against the 20 square meter room area and against the footprint of its nearest neighbours
    PASS IF: no single object dominates the room disproportionately; no object is so small relative to its neighbours that it reads as decorative rather than functional
    IF FAIL: flag the disproportionate object for replacement with an appropriately scaled alternative; do not resize in place
    EXCEPTIONS: NONE

S4  RULE: the length of the sofa or primary seating unit is proportional to the wall or zone it faces
    APPLY TO: the sofa or sofa bed in the lounge cluster; measure its length against the width of the wall or zone face it is oriented toward (the screen wall)
    PASS IF: sofa length is neither so long it crowds the flanking space nor so short it leaves the facing wall visually empty; sofa length must also be sufficient to seat at least three of the two to four expected guests with the host occupying the fourth seat or an occasional chair
    IF FAIL: flag the sofa for replacement with an appropriately sized unit; or reposition the sofa to face a wall of more proportionate width
    EXCEPTIONS: NONE


TIER 3 - HIGH: Activity Support
---------------------------------

A1 - Desk work with dual monitors and task lighting
-------------------------------------

A1  RULE: the desk work zone has enough clear floor area for the activity to be performed comfortably by a standard adult seated at a desk
    APPLY TO: the desk work zone; the activity pose is a seated adult at the desk with the office chair pulled to working position and reaching for both monitors and the task lamp; the activity pose is checked against the computed free-floor polygon
    PASS IF: the free-floor polygon within the zone contains the seated and chair-pulled-out body pose without collision
    IF FAIL: move furniture at the perimeter of the zone outward to expand the free-floor polygon until the activity pose fits within the free-floor polygon
    EXCEPTIONS: NONE

A1  RULE: the desk work zone is accessible from the main circulation path
    APPLY TO: the entry point of the desk work zone
    PASS IF: a clear path exists from the main circulation path to the desk zone entry point; path is wide enough for a standard adult
    IF FAIL: move the object blocking the desk zone entry point to restore the access path
    EXCEPTIONS: NONE

A1  RULE: the desk work zone is permanently clear and is not encroached on by movie viewing or sleeping furniture in any room mode
    APPLY TO: every object whose footprint or articulation might cross the desk work zone in lounge mode or in deployed-sleeping mode
    PASS IF: in every room mode the desk, office chair pullout zone, and task lamp remain unobstructed; no sofa bed or fold-out bed deployment footprint, no guest seating, and no occasional chair encroaches on the desk work zone
    IF FAIL: relocate the encroaching object to a position that does not cross the desk work zone in any mode; if no such position exists, flag the room as overloaded and recommend reducing furniture count or replacing pieces with smaller equivalents
    EXCEPTIONS: NONE


A1 - Movie watching with two to four friends
-------------------------------------

A1  RULE: the movie viewing zone has enough clear floor area for two to four guests to be seated facing the screen comfortably
    APPLY TO: the movie viewing and lounge zone; the activity pose is two to four seated adults distributed across the sofa and occasional seating with feet on the floor or on a coffee table; the activity pose is checked against the computed free-floor polygon
    PASS IF: the free-floor polygon within the zone contains all seated body poses without collision and provides foot space in front of each seat
    IF FAIL: move furniture at the perimeter of the zone outward to expand the free-floor polygon until all seated activity poses fit within the free-floor polygon; if no expansion is possible, reduce the occasional seating count to the maximum that fits
    EXCEPTIONS: NONE

A1  RULE: the movie viewing zone is accessible from the main circulation path
    APPLY TO: the entry point of the movie viewing zone
    PASS IF: a clear path exists from the main circulation path to the front of the sofa and to each occasional seat; path is wide enough for a standard adult and allows multiple guests to enter and seat themselves without queueing through a single bottleneck
    IF FAIL: move the object blocking the zone entry or any individual seat approach to restore the access path
    EXCEPTIONS: NONE

A1  RULE: if the movie viewing zone is also used as the overnight sleeping zone, the furniture that must be moved or unfolded for the transition can be repositioned without violating S1 or S2 in either configuration
    APPLY TO: every object involved in the lounge-to-sleep transition (sofa bed unfold motion, coffee table relocation, occasional chair stowage)
    PASS IF: each object has a valid resting position in both the lounge configuration and the deployed sleeping configuration that satisfies S1 and S2
    IF FAIL: identify an alternative resting position for the object that satisfies S1 and S2 in both configurations; if none exists, replace the sofa with a sofa bed of a smaller deployed footprint, or move the coffee table to a permanent stowed position along a free wall
    EXCEPTIONS: NONE


A1 - Overnight sleeping for guests
-------------------------------------

A1  RULE: the overnight sleeping zone has enough clear floor area for one adult sleeping surface in deployed configuration
    APPLY TO: the overnight sleeping zone in its deployed configuration; the activity pose is a standard adult lying full-length on the sleeping surface with at least one long side reachable on foot; the activity pose is checked against the computed free-floor polygon in the deployed configuration
    PASS IF: the deployed sleeping surface fits inside the room without violating S1 boundary or pairwise overlap; the free-floor polygon along at least one long side accommodates a standing adult
    IF FAIL: move furniture at the perimeter of the sleeping zone outward in the deployed configuration to expand the free-floor polygon; if the room cannot accommodate the deployed footprint, replace the sleeping surface with a narrower or shorter unit
    EXCEPTIONS: NONE

A1  RULE: the overnight sleeping zone is accessible from the main circulation path and from the entry door
    APPLY TO: the entry point of the deployed sleeping zone
    PASS IF: a clear path exists from the main circulation path to the side of the deployed sleeping surface; path is wide enough for a standard adult
    IF FAIL: move the object blocking the sleep zone entry to restore the access path
    EXCEPTIONS: NONE

A1  RULE: the lounge-mode furniture that must be moved or unfolded to create the overnight sleeping zone can be repositioned without violating S1 or S2 in the cleared configuration
    APPLY TO: every object that must be moved or stowed to deploy the overnight sleeping zone (coffee table, occasional chairs, rug edges, sofa cushions in sofa-bed mode)
    PASS IF: each object has a valid resting position in the cleared configuration that satisfies S1 and S2
    IF FAIL: identify an alternative resting position for the object that satisfies S1 and S2; if none exists, flag the sleeping zone as requiring a permanently smaller default lounge footprint
    EXCEPTIONS: NONE


GROUP: Zone Transformation
Priority: A1 > A4
-------------------------------------

A4  RULE: when the room transitions between office mode, movie mode, and overnight sleeping mode, every object that must be repositioned can move to a valid new position without violating S1, S2, or A1 in the transformed state
    APPLY TO: every object involved in any mode transformation between the three activity modes (office chair pushed under desk for movie mode; coffee table relocated for sleep deployment; sofa bed unfolded for sleep mode; occasional chairs stowed for sleep mode; rug edge cleared for sofa bed deployment)
    PASS IF: after repositioning, S1 and S2 rules pass for all objects in the scene in every mode; the activity pose fits in the transformed layout for the active mode
    IF FAIL: adjust the default position of the object so that its transformation path does not cause S1 or S2 violations; or replace the object with a version that folds or stacks to a smaller footprint
    EXCEPTIONS: NONE

A4  RULE: rugs or mats that must be rolled back or cleared for the sofa bed deployment can be rolled without moving furniture that sits on their edges
    APPLY TO: every rug in the scene that is used under furniture in lounge mode and must also be clearable to allow the sofa bed to unfold
    PASS IF: the rug can be rolled or cleared on the deployment side without requiring the repositioning of furniture that rests on its edges
    IF FAIL: move the furniture off the rug edges in the default lounge layout so the rug is independently rollable; recheck S1 overlap after adjustment
    EXCEPTIONS: NONE


TIER 4 - HIGH: Ergonomic Fit
------------------------------

GROUP: Desk Work Access
Priority: E2 > E1
-------------------------------------

E2  RULE: every door, drawer, and openable panel has a clear sweep zone in front of it
    APPLY TO: every hinged door, drawer, cabinet door, and openable panel in the scene JSON, including any drawer unit or filing cabinet near the desk; the executor identifies these by object type from the scene description
    PASS IF: articulation zone polygon intersects no other object's bounding box
    IF FAIL: move the object encroaching on the articulation zone away from the openable element until the zone is clear
    EXCEPTIONS: NONE

E2  RULE: the office chair at the desk has a clear pullout zone behind it
    APPLY TO: every chair, stool, and seat in the scene JSON that is pushed in against a desk or table surface, including the office chair at the desk
    PASS IF: chair pullout zone intersects no other object's bounding box
    IF FAIL: move the object behind the chair away until the pullout zone is clear; if the room does not allow this, flag the seating count as excessive for the available space
    EXCEPTIONS: NONE

E2  RULE: no companion object is placed within another object's access zone or articulation zone
    APPLY TO: every pair of objects in the scene JSON where one has an access zone and the other is a companion placed nearby (task lamp base inside the chair pullout zone; office storage drawer face within the desk chair pullout zone; coffee table within the sofa front access zone)
    PASS IF: no companion object bounding box overlaps any other object's access zone polygon
    IF FAIL: move the companion object out of the access zone; prefer moving it laterally along the wall rather than further into the room
    EXCEPTIONS: NONE


GROUP: Lounge and Guest Sleep Access
Priority: E2 > E1
-------------------------------------

E2  RULE: every openable element near the lounge and guest sleep cluster has a clear sweep zone in front of it
    APPLY TO: every hinged door, drawer, cabinet door, sofa bed unfold mechanism, and storage panel in or adjacent to the lounge cluster
    PASS IF: articulation zone polygon intersects no other object's bounding box in both lounge and deployed-sleep configurations
    IF FAIL: move the object encroaching on the articulation zone away from the openable element until the zone is clear in both configurations
    EXCEPTIONS: NONE

E2  RULE: every occasional seat and the sofa has a clear front access zone for sitting and standing
    APPLY TO: every chair, stool, and seat in the lounge cluster including the sofa and any occasional armchairs
    PASS IF: the seat's front access zone polygon intersects no other object's bounding box
    IF FAIL: move the object encroaching on the front access zone away from the seat; if the coffee table is the encroaching object, move it forward away from the sofa until the zone is clear
    EXCEPTIONS: NONE

E2  RULE: no companion object is placed within another object's access zone or articulation zone in the lounge cluster
    APPLY TO: every pair of objects in the lounge cluster where one has an access zone and the other is a companion placed nearby (coffee table inside sofa front access zone; floor cushion inside occasional chair pullout zone; rug edge inside sofa bed deployment zone)
    PASS IF: no companion object bounding box overlaps any other object's access zone polygon
    IF FAIL: move the companion object out of the access zone; prefer moving it laterally rather than further into the room
    EXCEPTIONS: NONE


STANDALONE
-------------------------------------

E3  RULE: every storage object is reachable by a standard adult without a step stool
    APPLY TO: every storage object in the scene JSON (office shelves, drawer units, filing cabinet, lounge sideboard or shelf, bedding storage); persona is matched to the user profile, a standard able-bodied adult
    PASS IF: a reach check confirms that the highest and lowest storage positions are within the reach envelope of a standard adult
    IF FAIL: lower the shelf or cabinet to bring the top shelf within reach; or raise the bottom drawer above knee level; flag if structural constraints prevent adjustment
    EXCEPTIONS: NONE

E3  RULE: the number of storage units currently specified is sufficient for the stated storage load; if load exceeds capacity, additional units must be added before layout is finalised
    APPLY TO: every named storage cluster in the scene (office storage near the desk and bedding-plus-lounge storage near the sofa); the executor sums available storage volume across all units in each cluster and compares against the estimated load derived from daily office work supplies and from bedding for two to four overnight guests on weekend evenings
    PASS IF: total storage volume in each cluster meets or exceeds the estimated load for that cluster
    IF FAIL: add one or more storage units of the same type to the cluster; recheck S3 storage adequacy and S2 wall-mount validity for each added unit
    EXCEPTIONS: NONE

E4  RULE: every seat and work surface is dimensioned for a standard adult's body and posture
    APPLY TO: every chair, stool, sofa, sofa bed, and desk in the scene JSON; persona matched to the user profile, a standard able-bodied adult
    PASS IF: seat height is appropriate for a standard adult to sit and stand from comfortably; legroom under the desk is sufficient for a standard adult's leg length
    IF FAIL: flag the object for replacement with a correctly sized alternative; do not resize in place
    EXCEPTIONS: NONE

E4  RULE: every screen or monitor is positioned at a comfortable viewing angle for a standard adult at the seated or standing position used for that screen
    APPLY TO: every monitor, television, and projection screen in the scene JSON; eye_pos at the desk is the seated office position for the dual monitors; eye_pos at the lounge is the seated sofa or occasional-seat position for the movie screen
    PASS IF: vertical and horizontal viewing angles are within a comfortable range for a standard adult at each eye_pos; no extreme neck flexion or rotation required between the two monitors at the desk or to the movie screen at the sofa
    IF FAIL: adjust screen height or tilt; or move the desk or sofa position to bring the viewing angle within range; recheck E2 access zones after move
    EXCEPTIONS: NONE


TIER 5 - MEDIUM
-----------------

STANDALONE
-------------------------------------

S5  RULE: the room reads as visually balanced when viewed from above
    APPLY TO: the full room layout; the executor renders the top-down view and passes it to VLM for balance and focal-point assessment
    PASS IF: VLM confirms no extreme visual weight imbalance between room quadrants; visual balance score is within an acceptable range for a multi-function spare room
    IF FAIL: move heavy visual elements (sofa, screen wall cluster, dense shelving) to counterbalance the dominant quadrant; recheck S1 after any move
    EXCEPTIONS: open-plan rooms and rooms with strong asymmetric architectural features (bay window, alcove on one side) may have a lower symmetry threshold - VLM assessment takes precedence over the balance score

A3  RULE: the workflow sequence for each multi-step activity follows a logical spatial path with minimal backtracking
    APPLY TO: every activity from Step 1B that has a defined sequence of object interactions (office mode: entry door to desk to office chair to monitors and task lamp; movie mode: entry door to sofa to screen and back, with detour to coffee table for snacks; sleep deployment: clear coffee table, unfold sofa bed, retrieve bedding from storage); steps[] is the ordered list of objects touched during each activity
    PASS IF: no backtracking or cross-path collision is found between steps in any activity; total path length is reasonable for a 20 square meter room
    IF FAIL: rearrange the objects in the sequence to reduce backtracking; prefer a linear or U-shaped path order; recheck E1 circulation and S1 overlap after rearrangement
    EXCEPTIONS: NONE

N2  RULE: no screen or sustained visual-work surface receives direct glare from a window or artificial light source
    APPLY TO: every monitor, television, projection screen, and reading position in the scene; eye_pos is the user's position at each (seated at desk for the two monitors; seated on sofa or occasional seat for the movie screen)
    PASS IF: the angle check confirms no window is in the direct forward arc of each screen normal; VLM glare check finds no visible light source in the point-of-view render from each eye_pos; the task lamp at the desk does not cast direct glare onto either monitor surface from the user's viewpoint
    IF FAIL: rotate the screen or desk to move the window out of the forward arc; or add a curtain or blind to the window causing glare; reposition the task lamp so its beam does not cross either monitor face; recheck S2 wall-mount validity if any blind mount is added
    EXCEPTIONS: NONE

N3  RULE: noise-generating zones are acoustically separated from quiet zones by distance or mass obstruction
    APPLY TO: every pair of zones in this room where one is a noise source and the other is a quiet zone; the movie viewing cluster is the noise source in movie mode, and the desk work zone is the quiet zone in office mode; the executor checks separation between these two clusters
    PASS IF: separation distance between the desk and the screen-sofa axis is sufficient that the desk is not directly in front of the movie speakers, or a mass obstruction (sofa back, shelving unit) sits on the direct path between the screen and the desk
    IF FAIL: move a mass object (sofa back, bookshelf, storage cabinet) onto the direct path between the screen and the desk; or increase separation distance by repositioning either cluster
    EXCEPTIONS: NONE


TIER 6 - LOW
-------------

STANDALONE
-------------------------------------

N1  RULE: priority activity zones have reasonable proximity to natural light
    APPLY TO: every activity zone from Step 1B that benefits from natural light; priority order in this room: desk work > movie viewing and lounge > overnight sleeping
    PASS IF: the desk work zone centroid is within an acceptable distance of a window so that daily work benefits from daylight; VLM brightness check confirms adequate daylight at the desk during typical weekday work hours based on sun path analysis
    IF FAIL: move the desk cluster toward the nearest window so daylight reaches the work surface; recheck E1 circulation and N2 glare after any move
    EXCEPTIONS: the overnight sleeping zone does not require window proximity - it may be positioned away from windows intentionally for blackout purposes; the movie viewing zone may also be positioned away from windows to reduce screen glare

N4  RULE: no object's bounding box overlaps the clearance zone of an HVAC vent, radiator, or heat source
    APPLY TO: every object near a known HVAC vent, radiator, or fixed heat source identified in the scene description; if no heat source is identified, flag the room as needing manual confirmation of vent and radiator locations before final placement
    PASS IF: the clearance check confirms no bounding box overlaps the vent clearance zone; the proximity check confirms no sleep or sustained-work surface is within the exclusion radius of a heat source
    IF FAIL: move the encroaching object away from the vent or heat source until clearance is restored
    EXCEPTIONS: NONE

\end{lstlisting}

\subsection{Evaluation Tools}
We provide implementation details of tools used for evaluation. 

\begin{lstlisting}[style=python, frame=single]
def boundary_check(obj, room):
    """
    Check if object's bounding box lies within the outer wall polygon.
    Returns: bool
    """
 
def bbox_collision(obj_a, obj_b):
    """
    Compute AABB pairwise overlap ratio between two objects.
    Returns: float, overlap percentage (0 means no collision)
    """
 
def contact_check(obj, surfaces):
    """
    Check if object is in contact with its required surface (floor/wall/ceiling).
    Returns: bool
    """
 
def wall_angle_check(obj, walls):
    """
    Compute angle between object facing and nearest wall normal.
    Returns: float, angle in degrees
    """
 
def object_exist(name, scene):
    """
    Check if a named object is present in the scene.
    Returns: bool
    """
 
def object_info(name, scene):
    """
    Retrieve geometric state of a named object.
    Returns: dict {dimensions [l,w,h], location [x,y,z], facing (degrees)}
    """
 
def size_ratio(obj, room):
    """
    Compute object footprint as a percentage of total room floor area.
    Returns: float, percentage
    """
 
def size_check(obj):
    """
    LLM judgement on whether object's absolute size is plausible for its type.
    Returns: str, one of {normal, big, small}
    """
 
def visual_balance_check(scene):
    """
    VLM judgement on room's visual balance from a top-down rendered view.
    Returns: str, free-text assessment of balance, focal point, and alignment
    """
\end{lstlisting}

\begin{lstlisting}[style=python, frame=single]
def pathfinding(start, end, scene, resolution=0.05):
    """
    Find shortest navigable path between two floor positions using A*.
    Args: start, end -- (x, z) world coordinates in metres
    Returns: list of (x, z) waypoints, or None if no path exists
    """
 
def path_width(path, scene):
    """
    Compute minimum navigable clearance along a given path.
    Returns: dict {min_width (m), bottleneck (x, z)}
    """
 
def articulation_zone(obj, scene):
    """
    Compute minimum clearance within the swing arc of a door or drawer.
    Returns: float, minimum clear distance in metres (0 means blocked)
    """
 
def chair_clearance(chair, table, scene):
    """
    Measure front and rear clearance distances for a chair.
    Returns: dict {front_clearance (m), rear_clearance (m)}
    """
 
def reach_check(persona, obj):
    """
    LLM judgement on reachability given user attributes and object_info().
    Returns: str, one of {reachable, too_high, too_low, too_deep}
    """
 
def posture_check(persona, obj):
    """
    LLM judgement on posture given user attributes and object dimensions.
    Returns: str, one of {fits, too_tight, poor_posture}
    """
\end{lstlisting}

\begin{lstlisting}[style=python, frame=single]
def free_floor_area(zone, scene):
    """
    Compute unoccupied floor area within a zone by subtracting object footprints.
    Returns: float, free area in m^2
    """
 
def object_in_zone(zone, activity, scene):
    """
    LLM judgement on whether activity-relevant objects are correctly placed in zone.
    Returns: str, one of {correctly_placed, misplaced, missing}
    """
 
def activity_support_check(activity, objects):
    """
    LLM judgement on whether objects can adequately support the required activity.
    Returns: str, one of {supported, unsupported, undersized, oversized}
    """
 
def inbetween_check(point_a, point_b, scene):
    """
    LLM judgement on sightline obstruction between two points using object_info().
    Returns: dict {status: clear/blocked, blocker_id: str or None}
    """
 
def total_path_length(activity_sequence, scene):
    """
    Compute cumulative travel distance across an ordered activity sequence.
    Returns: float, total path length in metres
    """
 
def workflow_check(activity_sequence, scene):
    """
    LLM judgement on whether object arrangement follows a logical workflow order.
    Returns: str, one of {optimal, suboptimal, backtracking, cross_path}
    """
 
def multi_activity_check(activities, scene):
    """
    LLM judgement on whether space supports multiple activities simultaneously.
    Reuses free_floor_area(), object_in_zone(), activity_support_check() per activity.
    Returns: dict {status: compatible/conflict, conflicting_pair: list or []}
    """
\end{lstlisting}

\begin{lstlisting}[style=python, frame=single]
def window_obs_ratio(window, scene, radius):
    """
    Compute per-object blocking ratio for objects within proximity of a window.
    Returns: list of dict {id: str, blocking_ratio: float}, sorted descending
    """
 
def screen_window_info(screen, window):
    """
    Compute geometric relationship between a screen and a window.
    Returns: dict {angle (degrees), distance (m)}
    """
 
def glare_check(screen, window):
    """
    LLM judgement on glare risk based on screen-window angle and distance.
    Returns: str, one of {no_risk, glare_risk}
    """
 
def zone_distance(zone_a, zone_b):
    """
    Compute Euclidean distance between two zone centroids.
    Returns: float, distance in metres
    """
 
def acoustic_check(zone_a, zone_b, scene):
    """
    LLM judgement on acoustic separation risk between a noise and quiet zone.
    Reuses activity_support_check() to verify sound-blocking objects on path.
    Returns: str, one of {separated, acoustic_risk}
    """
 
def vent_obs_ratio(vent, scene, radius):
    """
    Compute per-object blocking ratio for objects within proximity of a vent.
    Returns: list of dict {id: str, blocking_ratio: float}, sorted descending
    """
 
def distance_check(obj, heat_source, min_distance=0.5):
    """
    Check that a sensitive object maintains safe clearance from a heat source.
    Returns: bool
    """
\end{lstlisting}

\end{document}